\documentclass{SCIS2025}

\usepackage{microtype}
\usepackage{graphicx}

\usepackage{booktabs} 
\usepackage{hyperref}

\usepackage{amsmath}
\usepackage{amssymb}
\usepackage{mathtools}
\usepackage{amsthm}

\usepackage[capitalize,noabbrev]{cleveref}

\theoremstyle{plain}
\theoremstyle{definition}
\theoremstyle{remark}

\usepackage{multirow}
\usepackage{comment}
\usepackage{enumitem}
\usepackage{subfig}
\usepackage{adjustbox}

\newcommand{\model}{MG-LLM }
\newcommand{\modelsnosp}{MG-LLM}

\newcommand{\etc}{\textit{ etc}}
\newcommand{\etal}{\textit{ et al.}}
\usepackage{tabularx}
\usepackage[normalem]{ulem}
\useunder{\uline}{\ul}{}
\newcommand{\todo}[1]{%
    \ifnum#1>0%
        todo, todo, todo, todo, todo, todo, todo, todo, todo, todo, \todo{\numexpr#1-1\relax}%
    \fi%
}

\usepackage{amsmath,amsthm,amssymb}
\usepackage{graphicx}
\definecolor{ultramarine}{RGB}{0,0,0}
\newcommand{\rev}[1]{\textcolor{ultramarine}{#1}}

\begin{document}
\ArticleType{POSITION PAPER}
\Year{2025}
\Month{October}
\Vol{68}
\No{1}
\DOI{}
\ArtNo{}
\ReceiveDate{}
\ReviseDate{}
\AcceptDate{}
\OnlineDate{}
\AuthorMark{Xin Wang et.al.}
\AuthorCitation{Xin Wang, Zeyang Zhang, Linxin Xiao, Haibo Chen, Chendi Ge, Wenwu Zhu}
\title{Towards multimodal graph large language model}{Towards multimodal graph large language model}

\author[1,2]{Xin WANG}{}
\author[1]{Zeyang ZHANG}{}
\author[1]{Linxin XIAO}{}
\author[1]{Haibo CHEN}{}
\author[1]{Chendi GE}{}
\author[1,2]{Wenwu ZHU}{{wwzhu@tsinghua.edu.cn}}

\address[1]{Department of Computer Science and Technology, Beijing 100084, China}
\address[2]{Beijing National Research Center for Information Science and Technology, Beijing 100084, China}

\abstract{Multi-modal graphs, which integrate diverse multi-modal features and relations, are ubiquitous in real-world applications. 
However, existing multi-modal graph learning methods are typically trained from scratch for specific graph data and tasks, failing to generalize across various multi-modal graph data and tasks. 
To bridge this gap, we explore the potential of \textit{Multi-modal Graph Large Language Models (\modelsnosp)} to unify and generalize across diverse multi-modal graph data and tasks. We propose a unified framework of \textit{multi-modal graph data, task, and model}, discovering the inherent \textit{multi-granularity} and \textit{multi-scale} characteristics in multi-modal graphs. Specifically, we present five key desired characteristics for \modelsnosp: 1) unified space for multi-modal structures and attributes, 2) capability of handling diverse multi-modal graph tasks, 3) multi-modal graph in-context learning, 4) multi-modal graph interaction with natural language, and 5) multi-modal graph reasoning. We then elaborate on the key challenges, review existing literature, and highlight promising future research directions towards realizing these ambitious characteristics. Finally, we summarize existing multi-modal graph datasets pertinent for model training. We believe this paper can contribute to the ongoing advancement of the research towards \model for generalization across multi-modal graph data and tasks.}


\maketitle

\section{Introduction}
Multi-modal graphs, which integrate features from diverse modalities such as text, image, audio, and video, as well as capture the complex intra-model and inter-modal relations, are becoming increasingly ubiquitous in real-world applications. From social networks~\cite{bhattacharyya2024heterogeneous} and e-commerce~\cite{wang2023fashionklip} platforms to scientific discovery in biomedicine and materials science~\cite{marini2024multimodal, ye2024construction}, these complex data structures offer a richer, more holistic representation of interconnected entities than traditional unimodal graphs, which unlock more opportunities for advanced analytics, reasoning, and generation capabilities over multimodal information.

\begin{figure}
    \centering
    \includegraphics[width=1\textwidth]{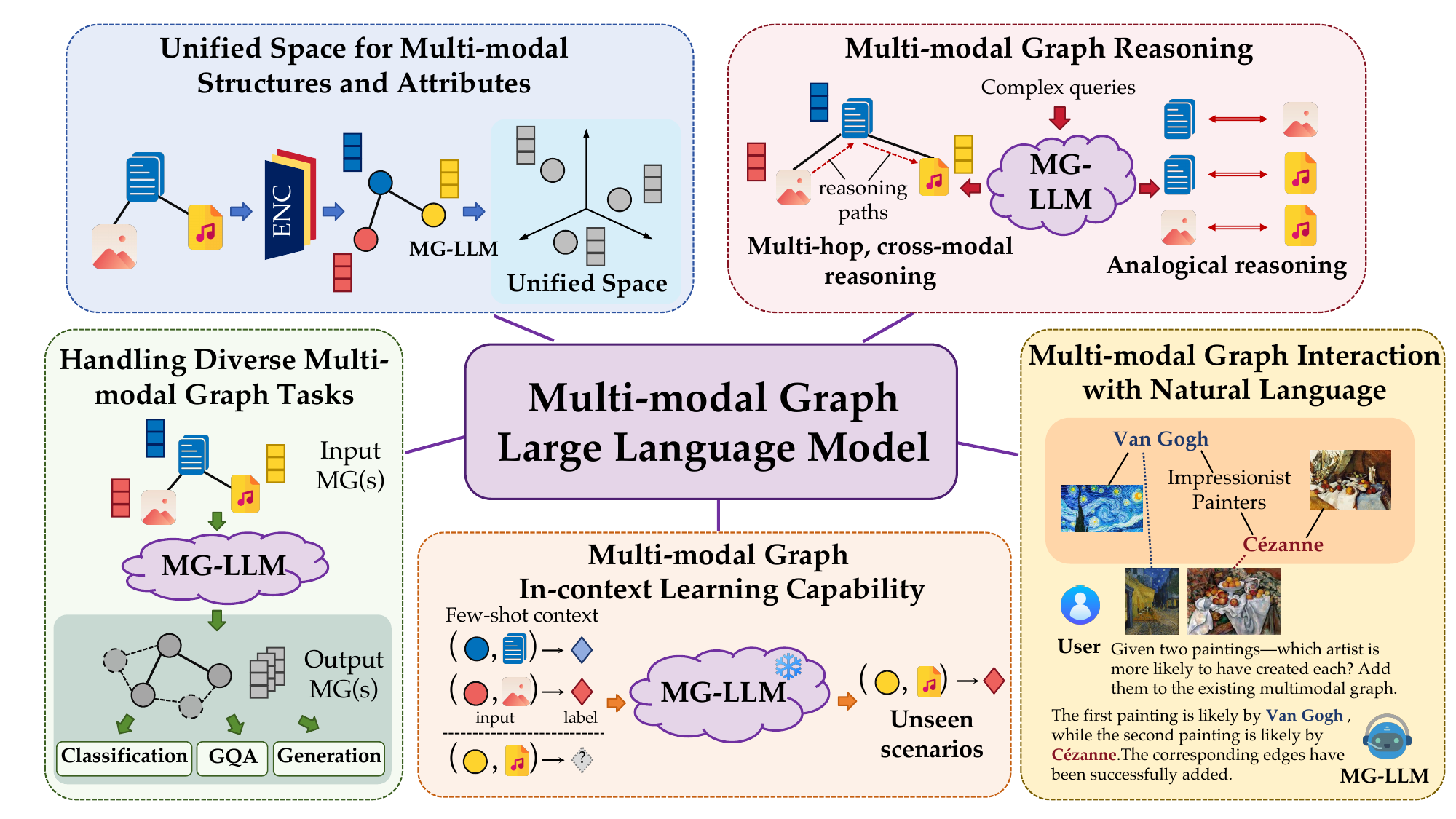}
    \caption{Key characteristics towards Multi-modal Graph Large Language Models (\modelsnosp).
    }
    \label{fig:characteristic}
\end{figure}

However, multi-modal graph learning currently faces a significant issue: existing methods are predominantly designed for specific tasks on particular types of graphs. This specialization often limits their applicability, preventing them from generalizing effectively across the vast diversity of multi-modal graph data and tasks encountered in practice. This lack of universality necessitates constant redesign and retraining for new scenarios, hindering the development of truly versatile and scalable solutions.

To bridge this gap, we explore the potential of Multi-modal Graph Large Language Models (\modelsnosp). Inspired by the remarkable success of large language models (LLMs) in unifying diverse natural language tasks, we propose that \model can serve as a powerful paradigm to unify and generalize across the complex landscape of multi-modal graph data and tasks. Our exploration begins by establishing a unified framework for multi-modal graph data, tasks, and models, which uncovers the inherent characteristics of multi-modal graphs, i.e., multi-granularity and multi-scale.

Specifically, we highlight that multi-modal graphs inherently exhibit multi-granularity, organizing information from fine-grained features like pixels and words to coarse-grained concepts such as entire images or documents, along with diverse structural complexities. This leads to multi-scale characteristics in multi-modal graph tasks, where inputs and outputs can vary dramatically in their scope, from individual nodes to entire graph structures.

Building upon this foundational understanding, we articulate five key desired characteristics towards \modelsnosp:
\begin{itemize}[leftmargin=0.5cm]
    \item Unified Space for Multi-modal Structures and Attributes: the ability to align and represent diverse multi-modal features and relations within a single unified embedding space, capable of handling highly irregular and continuous information.
    \item Ability of Handling Diverse Multi-modal Graph Tasks: the capacity to frame and solve all multi-modal graph tasks, from traditional discriminative problems like node classification to emerging generative tasks such as multi-modal content generation, under a unified generative modeling paradigm.
    \item Multi-modal Graph In-context Learning: the capability of performing novel tasks by leveraging a limited number of multi-modal graph examples provided directly within the prompt, without requiring explicit model fine-tuning.
    \item Multi-modal Graph Interaction with Natural Language: the possibility of enabling users to query, edit, generate, and reason about complex multi-modal graph-structured knowledge using intuitive natural language, bridging the gap between human language and structured data.
    \item Multi-modal Graph Reasoning: the proficiency in performing complex multi-hop, cross-modal reasoning, including analogical inference, by seamlessly combining information from various modalities and relational structures.
\end{itemize}

While the vision of \modelsnosp is ambitious, realizing these characteristics presents significant challenges, ranging from developing unified multi-modal graph vocabularies and tokenization schemes to multi-modal graph architectures capable of large-scale pretraining. This paper delves into these key challenges, reviews existing research that moves towards this paradigm, and outlines promising future research directions to accelerate the development of \model for generalizing across diverse multi-modal graph data and tasks. Finally, we summarize existing multi-modal graph datasets that could be useful for the training and evaluation of such models. This work aims to foster progress towards a new era of multi-modal graph intelligence.

Our main contributions are summarized as follows:
\begin{itemize}
    \item We explore the potential of \textit{Multi-modal Graph Large Language Models (\modelsnosp)} to unify and generalize across diverse multi-modal graph data and tasks, aiming for universal generalization across diverse multi-modal graph data and tasks. We systematically discuss the potential of \modelsnosp, for the first time, to the best of our knowledge.
    \item We present a unified framework for understanding multi-modal graph data, tasks, and models, highlighting their inherent multi-granularity and multi-scale characteristics for designing \modelsnosp.
    \item We propose five essential characteristics that \model should possess, coupled with detailed discussions of challenges and future directions, thereby setting a clear research roadmap. We also summarize existing multi-modal graph datasets and tasks for developing \modelsnosp.
\end{itemize}

The domain of multi-modal learning on graphs is rapidly advancing. \rev{We distinguish our work from several related lines of research. 1) Compared to surveys on multi-modal graph learning~\cite{ektefaie2023multimodal, peng2024learningmultimodalgraphssurvey}, which primarily catalog existing techniques, our work introduces a novel conceptual framework and a forward-looking vision for a unified MG-LLM. 2) Unlike existing GraphLLMs~\cite{chen2024llaga, kong2024gofa}, which mainly focus on adapting unimodal graph data for LLMs, we address the more complex challenge of natively handling graphs with rich multi-modal attributes. 3) Distinct from general-purpose Omni-MLLMs~\cite{jiang2025from}, we argue for a specialized paradigm that deeply integrates graph-structured reasoning rather than treating graphs as just another input modality.} By identifying the inherent characteristics of multi-modal graphs and the desired characteristics for \modelsnosp, we chart a new research roadmap towards developing multi-modal graph large language models, thereby complementing surveys of the existing state-of-the-art by looking towards a next-generation paradigm.

\begin{figure}[t]
    \centering
    \includegraphics[width=0.95\textwidth]{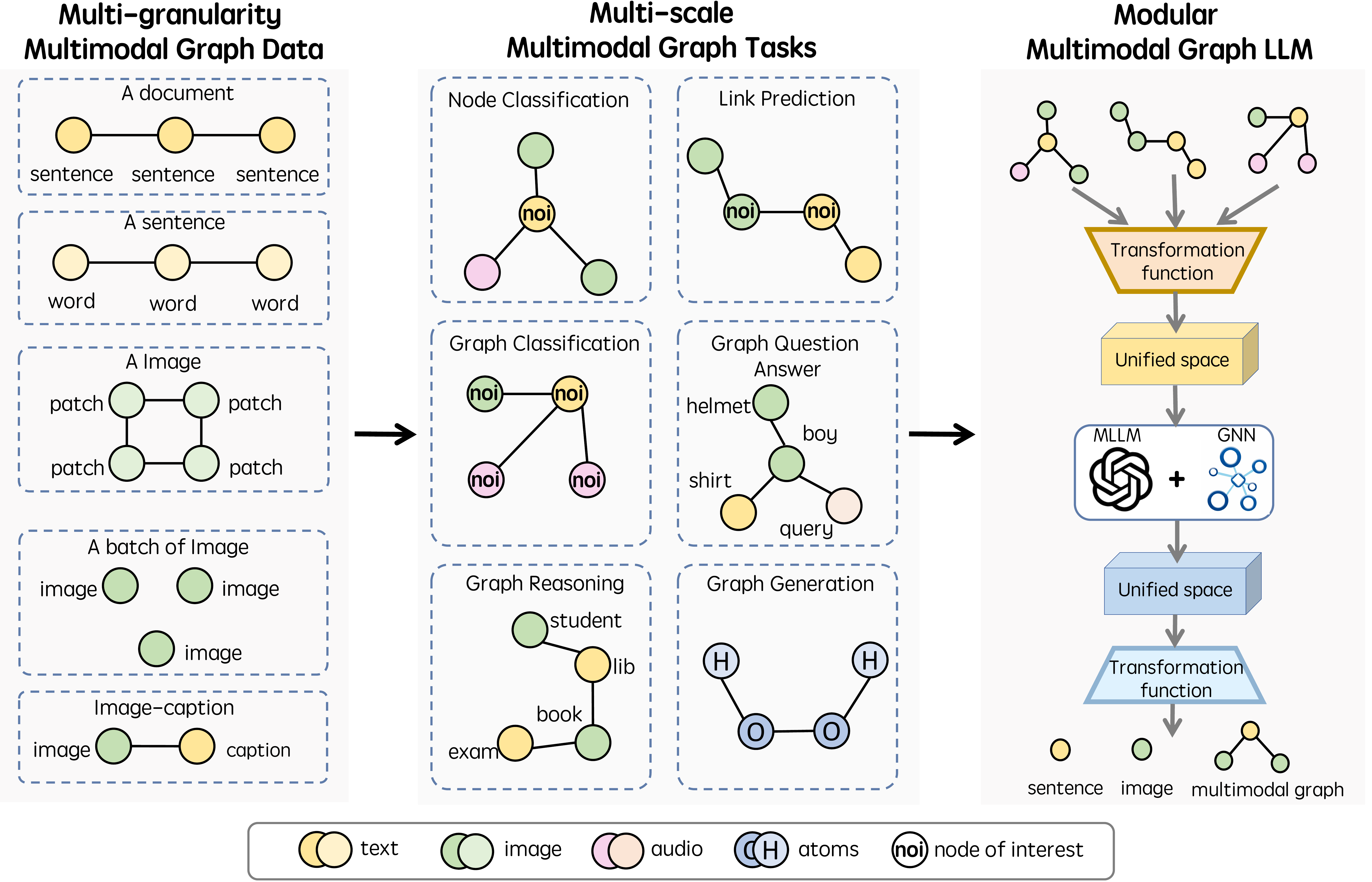}
    \caption{ Unified view of multimodal graph data, tasks, and model towards \modelsnosp.
    }
    \label{fig:framework}
\end{figure}

\section{Towards a unified view of multi-modal graph data, task, and model}

In this section, we introduce a unified framework of multi-modal graph data, task, and model, and remark on the inherent characteristics in multi-modal graphs, serving as a foundation to discuss the desired characteristics of \modelsnosp. The overall framework is shown in Figure~\ref{fig:framework}.

\subsection{Unified formulation of multi-modal graph data}
In this section, we define multi-modal graphs, extending standard graphs with diverse node and edge modalities. Then we outline three decomposable types (feature-, node-, graph-level), and their versatility in representing various data forms, from single instances to full datasets. Moreover, we give remarks on the challenges of {\it indecomposability} and {\it multi-granularity} for building effective \modelsnosp.  

\paragraph{Graph} A graph $\mathcal{G}=(\mathcal{V},\mathcal{E})$ consists of a finite set of vertices $\mathcal{V}=\{v_1,v_2,\dots,v_n\}$ and a set of edges $\mathcal{E}\subseteq\mathcal{V}\times\mathcal{V}$, each edge being an ordered pair of vertices denoting a directed relation between them.

\begin{figure}[t]
    \centering
    \includegraphics[width=0.6\textwidth]{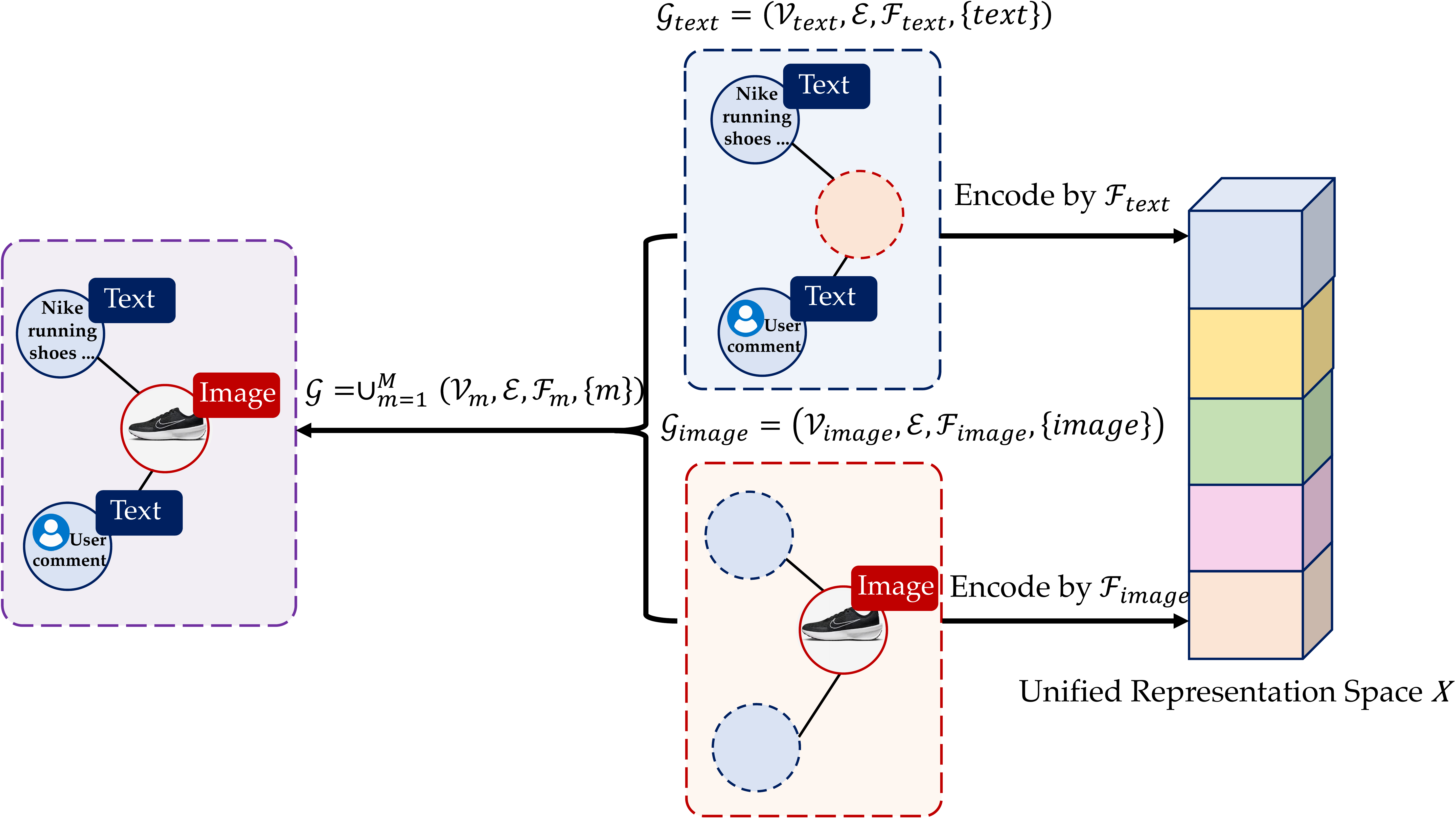}
    \caption{\rev{ An illustrative diagram and a running example of a multi-modal graph, take a small product graph with text and image nodes as an example.}
    }
    \label{fig:case}
\end{figure}

\paragraph{Multi-modal graph} A modality is a distinct type or source of information associated with nodes or edges. Let $\mathcal{M} = \{1, 2, \dots, M\}$ denote the set of all modalities. \rev{We define a mapping set \rev{ $\mathcal{F} = \{\mathcal{F}_m\}_{m=1}^M$} for all $M$ modalities. For each $m \in \mathcal{M}$, the mapping \rev{$\mathcal{F}_m$} maps from the modality-specific node feature space $\mathcal{V}$ to a shared representation space $\mathcal{X}$}, such that \rev{$\mathcal{F}_m: \mathcal{V} \to \mathcal{X}$}. The space $\mathcal{X}$ serves as a unified embedding space for all modalities. Similarly, we define a mapping \rev{$\mathcal{F}_m$} from the modality-specific edge feature space $\mathcal{E}$ to the shared representation space $\mathcal{X}$, such that $\mathcal{F}_m: \mathcal{E} \to \mathcal{X}$. 
For simplicity, we reuse \rev{$\mathcal{F}_m$} as the modality-$m$ map for both nodes and edges. 
We likewise leave out multimodal edges in the formulation, even though extending them would be straightforward. The features in different modalities could be texts, images, audios, videos, \etc. A multi-modal graph can be defined by the quadruple $\mathcal{G}=(\mathcal{V},\mathcal{E}, \mathcal{F}, \mathcal{M})$. \rev{To make the formulation more concrete, a running example is shown in Figure~\ref{fig:case}.}

\paragraph{Special cases of decomposable multi-modal graphs} 
By instantiation of the modality set and feature mapping, we could obtain several classic types of multi-modal graphs which are ubiquitous in real-world applications~\cite{peng2024learningmultimodalgraphssurvey}. These types of multi-modal graphs share the same assumption that they can be {\it decomposed} by modality from different perspectives, i.e., feature, node, and graph:
\begin{itemize}[leftmargin=0.5cm]
    \item Feature-level Multi-modal Graph, where the features of nodes or edges come from different modalities, i.e.,
    $
        \mathcal{G} = \bigcup_{m=1}^M \mathcal{G}_m = \bigcup_{m=1}^M (\mathcal{V},\mathcal{E},\mathcal{F}_m,\{m\}),
    $
    where $m$ representing one modality in $M$ modalities. The feature of node $v$ could be represented as $\mathbf{x}(v) =\bigoplus_{m=1}^M \mathcal{F}_{m}(v)$. For example, on an e-commerce product graph~\cite{zhu2025mosaicmodalitiescomprehensivebenchmark}, each node has the feature of product title and image, i.e., $\mathbf{x}(v) =\bigoplus ( \mathcal{F}_{{\text{text}}}(v), \mathcal{F}_{{\text{image}}}(v))$.
    
    \item Node-level Multi-modal Graph, where the nodes or edges come from different modalities, while each node or edge has unimodal features, i.e.,
    $
        \mathcal{G} = \bigcup_{m=1}^M \mathcal{G}_m = \bigcup_{m=1}^M (\mathcal{V}_m,\mathcal{E},\mathcal{F}_m,\{m\}),
    $
     where $m$ representing one modality in $M$ modalities, and $\mathcal{V}_i \bigcap \mathcal{V}_j = \emptyset$.
    For example, on a multimodal knowledge base \cite{10.1007/978-3-030-41407-8_9}, each node might be either an image or a textual description.
    \item Graph-level Multi-modal Graph, where the graphs come from different modalities, while each graph has unimodal features, i.e.,
     $
        \mathcal{G} = \bigcup_{m=1}^M \mathcal{G}_m = \bigcup_{m=1}^M (\mathcal{V}_m,\mathcal{E}_m,\mathcal{X}_m,\{m\}),
    $
    where $m$ representing one modality in $M$ modalities, and $\mathcal{E}_i \bigcap \mathcal{E}_j = \emptyset$.
    For example, on a multi-modal question answering graph \cite{8953451}, we may have a graph with images, a graph with texts, \etc.
\end{itemize}

\remark{\textbf{(Indecomposable characteristics)} Although practitioners may model their data with the aforementioned decomposable multi-modal graphs for convenience, most multi-modal graphs in real-world scenarios, with nodes and edges having features from various modalities, may not be easily decomposable to several uni-modal subgraphs. For instance, in a multi-modal graph where the text node says `The Transformer was incredible!', the image node shows Optimus Prime (a central robot character from the Transformers movie series), and the knowledge node links to the movie Transformers, only joint reasoning over all three nodes can resolve the ambiguity and correctly interpret `Transformer' as a film character rather than a neural network architecture. Due to the {\it indecomposable} characteristics, established multi-modal fusion techniques in other multi-modal fields~\cite{zhao2024deep} may fail to flexibly solve multi-modal graph problems, calling for the need of {\it native modeling of multi-modal graph data} in multi-modal graph large language models. }

\paragraph{Special cases of multi-modal graph instances} The versatility of multi-modal graphs allows them to represent not only complex inter-modal relationships but also instances or datasets composed of single or multiple modalities as special cases, e.g., instances of texts, images, audios, videos,\etc, or pairs of image-captions, text-audios,\etc. Here are examples of representing single-modal instances:
\begin{itemize}[leftmargin=0.5cm]
    \item a text sequence can be represented by a multi-modal graph with a single text-attributed node, i.e., \rev{$\mathcal{G} = \left(\mathcal{V}=\{v\}, \mathcal{E}=\emptyset, \{\mathcal{F}_{\text{text}}\}, \{\text{text}\}\right)$}, where \rev{$\mathcal{F}_{\text{text}}$} is the text feature mapping function.
    \item a text sequence, more granularly, can be represented by a multi-modal graph where each word is a node and sequential or semantic connections form edges, i.e., \rev{$\mathcal{G} = \left(\mathcal{V}, \mathcal{E}, \{\mathcal{F}_{\text{word}}\}, \{\text{word}\}\right)$}, where $\mathcal{V}=\{v_1, \dots, v_L\}, \mathcal{E}=\{(v_i, v_{i+1})\}_{i=1}^{L-1}$, $L$ is the length of the text sequence, and {$\mathcal{F}_{\text{word}}$} is the word feature mapping function.
    \item an image can be represented as a multi-modal graph where pixels are nodes and their grid-like inter-connections form edges, i.e., an $H \times W$ image can be \rev{$\mathcal{G} = \left(\mathcal{V}=\{v_{ij}\}_{i=1,j=1}^{H,W}, \mathcal{E}_{\text{grid}}, \{\mathcal{F}_{\text{pixel}}\}, \{\text{pixel}\}\right)$}, where $\mathcal{E}_{\text{grid}}$ represents grid-like inter-connections, and \rev{$\mathcal{F}_{\text{pixel}}$} is the pixel feature mapping function.
\end{itemize}
Beyond individual instances, multi-modal graphs can efficiently represent entire datasets. Here are some examples.
\begin{itemize}[leftmargin=0.5cm]
    \item A batch of images can be represented as a multi-modal graph with several image-attributed nodes without any edges, i.e., \rev{$\mathcal{G} = \left(\mathcal{V}=\{v_1, \dots, v_K\}, \mathcal{E}=\emptyset, \{\mathcal{F}_{\text{image}}\}, \{\text{image}\}\right)$}, where \rev{$\mathcal{F}_{\text{image}}$} is the image feature mapping function, and $K$ is the number of images.
    \item An image-captioning dataset can be represented as a multi-modal graph where edges connect image-attributed nodes to their corresponding text-attributed caption nodes, i.e., for $K$ image-caption pairs, \rev{$\mathcal{G} = \left(\mathcal{V}_{\text{image}} \cup \mathcal{V}_{\text{text}}, \mathcal{E}_{\text{image-text}}, \{\mathcal{F}_{\text{image}}, \mathcal{F}_{\text{text}}\}, \{\text{image, text}\}\right)$}, where $\mathcal{V}_{\text{image}} = \{v_1, \dots, v_K\}$ are image nodes, $\mathcal{V}_{\text{text}} = \{u_1, \dots, u_K\}$ are caption nodes, $\mathcal{E}_{\text{image-text}} = \{(v_k, u_k) \mid 1 \le k \le K\}$ are edges connecting images to their captions, and \rev{$\mathcal{F}_{\text{image}}$ and $\mathcal{F}_{\text{text}}$} are the image and text feature mapping functions, respectively.
\end{itemize}
This ability to abstract various data forms into a unified graph structure underscores the expressive power of multi-modal graphs.

\remark{\textbf{(Multi-granularity characteristics)} Multi-modal graphs inherently possess the ability to organize data with {\it multi-granularity} across modalities, features, and structures. However, this capability is a double-edged sword. While they can represent vast amounts of information, they also introduce significant challenges for models to process them flexibly. Unlike other domains that feature units of roughly uniform granularity, such as word tokens in natural language processing (NLP) or image pixels in computer vision (CV), multi-modal graphs often contain units ranging from fine-grained features (e.g., pixels, words) to coarse-grained concepts (e.g., full images, entire documents). To build an effective \model capable of flexibly handling information at diverse granularities on multi-modal graphs, it may be necessary to design a {\it unified multi-modal graph vocabulary and tokenizer} for learning multi-modal graph representations in a shared space.}

\subsection{Generative modeling of multi-modal graph tasks}
In this section, we can frame, through generative modeling, that \textit{all multi-modal graph tasks are multi-modal graph generation}. 
Due to the inherent {\it multi-granularity} characteristics of multi-modal graphs, we can unify several classical discriminative tasks and emerging generative tasks under a single {\it generative perspective}, which can bring advantages of unified task forms, types, and interfaces. Suppose \model learns a conditional probability distribution to generate an output multi-modal graph $\mathcal{G}_{\text{out}}$ given an input $\mathcal{G}_{\text{in}}$. Formally, the objective is to model:
\begin{equation}
 P(\mathcal{G}_{\text{out}} | \mathcal{G}_{\text{in}}; \Theta),
\end{equation}
where $\Theta$ are the \modelsnosp 's parameters. Various multi-modal graph tasks can be redefined generatively:

\begin{figure}[t]
    \centering
    \includegraphics[width=0.95\textwidth]{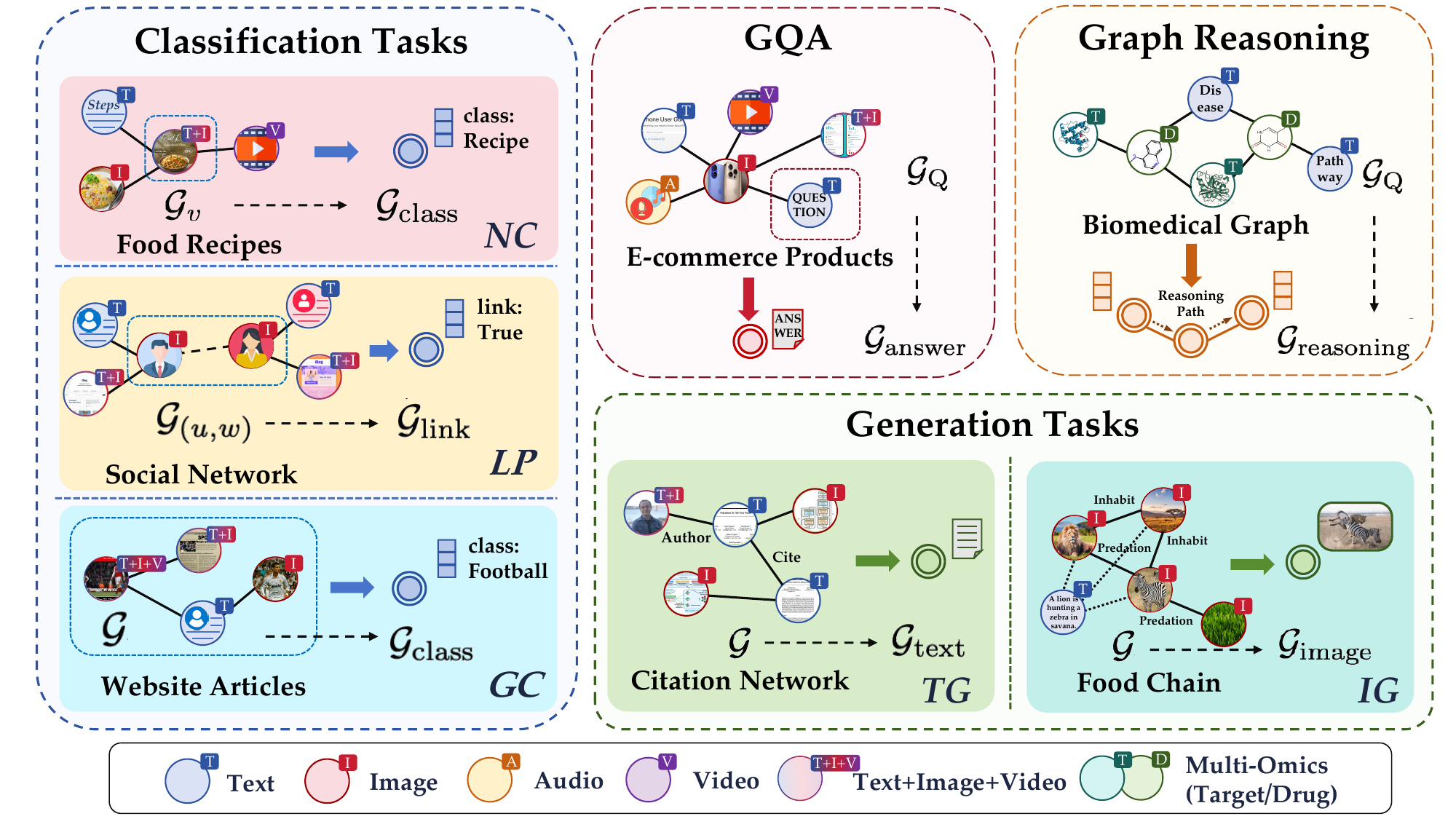}
    \caption{ \rev{Concrete working example of generative modeling across different tasks.}
    }
    \label{fig:generation}
\end{figure}

\begin{itemize}[leftmargin=0.5cm]
    \item \textbf{Multi-modal Node Classification (NC)} aims to take a multi-modal ego-graph centered around a target node as input, and generate a multi-modal graph representing the predicted class, i.e., $P(\mathcal{G}_{\text{class}} \mid \mathcal{G}_v)$, where $\mathcal{G}_v = (\mathcal{V}_v, \mathcal{E}_v, \mathcal{F}_v, \mathcal{M}_v)$ is a subgraph centered at node $v \in \mathcal{V}$ and  \rev{$\mathcal{G}_{\text{class}} = (\{v\}, \emptyset, \{\mathcal{F}_m\}, \{m\})$} is an output graph where $\mathcal{F}_m(v)$ encodes the class label (e.g., text or image).

    \item \textbf{Multi-modal Link Prediction (LP)} takes a multi-modal subgraph containing two endpoint nodes and their local neighborhood, and outputs a multi-modal graph indicating the link's existence or properties, i.e.,
    optimizing the objective $P(\mathcal{G}_{\text{link}} \mid \mathcal{G}_{(u,w)})$, where $\mathcal{G}_{(u,w)} = (\mathcal{V}_{(u,w)}, \mathcal{E}_{(u,w)}, \mathcal{F}_{(u,w)}, \mathcal{M}_{(u,w)})$ is a subgraph with nodes $u,w \in \mathcal{V}$ and their neighborhood, and  \rev{$\mathcal{G}_{\text{link}} = (\{v\}, \emptyset, \{\mathcal{F}_{(m)}\}, \{m\})$} is an output graph where $v$'s feature $\mathcal{F}_m(v)$ encodes link existence or type.

    \item \textbf{Multi-modal Graph Classification (GC)} takes the entire multi-modal graph as input and generates a multi-modal graph representing the graph's overall class or category, i.e., optimizing the objective $P(\mathcal{G}_{\text{class}} \mid \mathcal{G})$
    , where $\mathcal{G} = (\mathcal{V},\mathcal{E}, \mathcal{F}, \mathcal{M})$ is the input graph and \rev{ $\mathcal{G}_{\text{class}} = (\{v\}, \emptyset, \{\mathcal{F}_m\}, \{m\})$} is an output graph where $v$'s feature $\mathcal{F}_m(v)$ describes the predicted class.

    \item \textbf{Multi-modal Graph Question Answering (GQA)} aims to generate an answer based on a multi-modal graph $\mathcal{G}$ and a text-attributed query node $v_Q$, i.e., optimizing the objective $P(\mathcal{G}_{\text{answer}} \mid \mathcal{G}_{\text{Q}})$, where $\mathcal{G}_{\text{Q}} = (\mathcal{V} \cup \{v_Q\}, \mathcal{E} \cup \mathcal{E}_Q, \mathcal{X} \cup \{\mathcal{F}_{\text{text}}(v_Q)\}, \mathcal{M} \cup \{\text{text}\})$ is the graph augmented with $v_Q$ and potential edges $\mathcal{E}_Q$ and $\mathcal{G}_{\text{answer}}$ is the generated answer graph (e.g., a text/image node or a subgraph).

    \item \textbf{Multi-modal Graph Reasoning (GR)} extends GQA with complex multi-hop reasoning. The generated output $\mathcal{G}_{\text{reasoning}}$ may embody complex logical structures or a chain of thought, i.e., optimizing the objective 
    $P(\mathcal{G}_{\text{reasoning}} \mid \mathcal{G}_{\text{Q}}) $, where $\mathcal{G}_{\text{Q}}$ includes the graph and query, and $\mathcal{G}_{\text{reasoning}}$ encapsulates the reasoning result, which could be the thinking process like chain-of-thoughts or graph-of-thoughts.

    \item \textbf{Multi-modal Graph Text Generation (TG)} utilizes multi-modal graph information to generate coherent text sequences, i.e., optimizing the objective $P(\mathcal{G}_{\text{text}} \mid \mathcal{G})$, 
    where $\mathcal{G}_{\text{text}}$ is the generated text, such as a summary of a group of papers cited by each other or a new git patch based on correlated git commits.

    \item \textbf{Multi-modal Graph Image Generation (IG)} aims to generate novel images, where a multi-modal graph with textual descriptions, structured data, or other modal inputs can serve as a basis, i.e., optimizing the objective $P(\mathcal{G}_{\text{image}} | \mathcal{G}) $, where $P(\mathcal{G}_{\text{image}})$ is the generated image, such as a descriptive image of the food chain based on a multi-modal graph of ecosystems or a novel-style painting based on a multi-modal graph of artist networks. 
\end{itemize}
\rev{We provide concrete applications for the aforementioned tasks under a generative perspective, as illustrated in Figure~\ref{fig:generation}}. Further details on the datasets for these tasks are available in Section~\ref{sec:dataset}.
\remark{\textbf{(Multi-scale Characteristics)} This generative paradigm offers a powerful and flexible framework for modeling diverse multi-modal graph tasks with a unified interface. Since multi-modal graphs inherently represent \textit{multi-granular} information, ranging from unimodal instances and bi-modal instances to entire multi-modal datasets, the resulting input and output spaces are exceptionally versatile, accommodating a wide array of tasks. This versatility, however, introduces \textit{multi-scale characteristics} for multi-modal graph tasks: the input and output graphs can differ significantly in scope. For example, a task might take an entire graph as input but require only a single node or path as output (as in a GQA task), or vice versa for other potential tasks. This disparity in scales and granularity poses critical challenges for \textit{unified task modeling} and \textit{task prompt design} in \modelsnosp, as the scales and semantic levels of inputs and outputs can vary widely.}

\subsection{Unified view of multi-modal graph models}
In this section, we provide a unified view on current multi-modal graph models by first proposing a transformation function. Then, we bring together the two main categories that are moving towards \modelsnosp, and we will briefly overview these approaches.

\paragraph{Transformation function}
We first propose a transformation function, denoted as $\mathcal{T}$, which maps one multi-modal graph to another. This transformation often involves a reduction in the number of modalities and the size of the graph, thereby facilitating processing by models or aligning the output with desired modalities and formats. This transformation function $\mathcal{T}$ can be parameter-free (e.g., designed via heuristic rules) or parametric (learned using neural networks), i.e., $\mathcal{T}_\theta$ with the parameters $\theta$, which we omit for brevity.

For instance, in the input space, $\mathcal{T}$ can convert an entire multi-modal graph into text. This could involve image captioning for image features, speech-to-text conversion for audio features, and textualizing edges into XML-like languages for structures. Consequently, a complex multi-modal graph is transformed into a simplified multi-modal graph where nodes primarily possess text attributes. Formally, given an input multi-modal graph $\mathcal{G}$, the transformation $\mathcal{T}$ could yield $\mathcal{G}'=(\mathcal{V}',\mathcal{E}', \mathcal{F}', \{\text{text}\})$, where $\mathcal{F}'$ is the text feature mapping function.

Similarly, in the output space, $\mathcal{T}$ can summarize a multi-modal graph into a single label, a document, or an image. An image output, for example, might not solely originate from extracting a single image-attributed node but could also involve rendering an entire multi-modal graph into a coherent visual representation (e.g., visualizing a family tree).

In this view, a multi-modal graph model can be seen as first transforming the input multi-modal graph via a transformation, then modeling it, and finally transforming it again into the desired output. This can be formally expressed as:
\begin{equation}
\mathcal{G}_{\text{out}} = \mathcal{T}_{out} (\phi_\theta(\mathcal{T}_{int}(\mathcal{G}_{\text{in}}))),
\end{equation}
where $\mathcal{G}_{\text{in}}$ is the input multi-modal graph, $\mathcal{T}_{in}$ is the input transformation, $\phi_\theta$ is the core multi-modal graph model, $\mathcal{T}_{out}$ is the final transformation, and $\mathcal{G}_{\text{out}}$ is the generated output multi-modal graph. This generalized framework highlights the crucial role of these transformations in aligning diverse multi-modal graph data with the model's capabilities and desired output formats.

\paragraph{Multi-modal graph neural networks} The transformation function $\mathcal{T}$ is typically considered a parametric function. Here, information from each modality is initially mapped into a learned representation via trained modality-specific encoders. Subsequently, a GNN leverages its message-passing mechanism to learn from these representations and derive the required labels for downstream tasks. For instance, multi-modal graph convolution networks (MGCNs) \cite{Wei2019MMGCN, Song2023Multicenter, Zeng2023Hypergraph} utilize the learned multimodal representation to form an adjacency matrix, and multimodal graph attention networks (MGATs) fuse information from different modalities by assigning different attention weights to each node \cite{tian2022recipe2vecmultimodalreciperepresentation, he-wang-2023-multimodal, 9950299}. The primary advantage of these approaches lies in the MGNN's ability to explicitly utilize structural information within the graph. However, a significant drawback is the lack of flexible input and output spaces, which complicates the development of unified foundation models for multi-modal graphs. Furthermore, this late fusion function can lead to substantial information loss, making it challenging to capture fine-grained modal interactions.

\paragraph{Graph large language models} GraphLLMs often employ different strategies for the transformation function $\mathcal{T}$. 1) Some works utilize a non-parametric transformation function. For instance, they might describe the entire graph as text, which is then processed by an LLM~\cite{wang2024can,guo2023gpt4graph,zhao2023gimlet,liu2024git,liu2023evaluating,fatemi2023talk,hu2023beyond}. Alternatively, they might transform the graph into image-text pairs to be processed by a vision-language model (VLM). The advantage of these methods is their ability to leverage the flexible input and output spaces of LLMs or VLMs, enabling them to handle a wide range of tasks. However, they are heavily dependent on the inherent capabilities of the underlying LLM or VLM. Describing a complex graph entirely in text can lead to very long contexts, making comprehension difficult for the model. Moreover, certain modal information may be inherently challenging to textualize (e.g., an image converted to text can suffer significant information loss). 2) Other works employ a parametric transformation function. For instance, LLaGA \cite{chen2024llaga} employs a parametric projector to transform graph data into structured sequences, which are then embedded into the token space and fed into a large language model (LLM) for further processing. This mapping enables LLMs to effectively handle graph-structured data, enhancing their versatility, generalizability, and interpretability. GOFA \cite{kong2024gofa} interleaves randomly initialized graph neural network (GNN) layers within a frozen pre-trained LLM, organically combining semantic and structural modeling capabilities. This design leverages the GNN’s strength in processing graph structures alongside the LLM’s generative and reasoning abilities.
It is important to note that current GraphLLMs rarely address multi-modal graph problems directly.

\remark{\textbf{(Modular characteristics)} For building a \textit{native \modelsnosp}, the transformation function $\mathcal{T}$ might ideally be an identity mapping, i.e., the input multi-modal graph is directly fed into the primitive \model without any information loss. To achieve this ambitious goal, we might have to pretrain the model on extremely large multi-modal graphs, e.g., the entire internet, so that sufficient pairwise data across modalities could directly empower the model with comprehensive knowledge of various modalities and graph structures. However, this monolithic approach might be impractical in the near future due to the considerable computational expenses and data acquisition challenges. One possible solution to circumvent this limitation might be building a \textit{modular multi-modal graph LLM}, which integrates various parameterized modules designed for specific functions to understand as well as generate both the structures and multi-modal information within multi-modal graphs. This modularity could allow for more efficient training and flexible adaptation to diverse multi-modal graph tasks.}
\rev{It is important, however, to distinguish this from a generic Omni-MLLM with a graph \textit{plugin}, as an ideal modular MG-LLM would still feature deeply co-designed components.}

\section{Towards multi-modal graph large language models}
In this section, we will delve into the essential characteristics that a multi-modal graph large language model should possess, the core challenges in achieving these characteristics, current relevant approaches, and potential future research directions. The key characteristics are illustrated in Figure~\ref{fig:characteristic}.

\subsection{Unified space for multi-modal structures and attributes}
\paragraph{Desired characteristics}
\model is envisioned to effectively process information across diverse domains and a wide range of data modalities~\cite{peng2024learningmultimodalgraphssurvey}. Real-world applications in healthcare, finance, scientific discovery, and social media often present highly heterogeneous structures and data types~\cite{cai2024multimodalgraph}. Therefore, \model should possess strong domain transferability and the capacity for broad representation generalization. 
A key characteristic of \model is the ability to align diverse \textit{multi-modal features} and \textit{relations} within a truly unified representation space. This requires developing a comprehensive \textit{multi-modal graph vocabulary} capable of capturing highly irregular, multi-level, and continuous structural and attribute information. Such a space should facilitate the learning of \textit{transferable patterns} that generalize across heterogeneous data domains. The goal is to create a seamless integration where information from text, images, audio, video, and structured graph topologies can be jointly processed and understood. This unified space should minimize redundancy while maximizing information retention and the faithful representation of relational semantics, enabling robust domain transferability and flexible interpretation of various data inputs.

\paragraph{Key challenges}
One of the fundamental challenges in building \model is the inherent \textit{heterogeneity of data domains}. Data often originates from various sources with different formats and structures, such as biomedical data (e.g., proteins, drugs) \cite{zitnik2018modeling}, social networks (e.g., text, images) \cite{wu2020comprehensive}, and multi-modal knowledge graphs (e.g., language, images, temporal data) \cite{wang2021kepler}. These domains exhibit distinct properties, including categorical, continuous, or structured data types \cite{srivastava2014dropout}. This heterogeneity poses a significant obstacle in designing a unified model capable of efficiently integrating and processing such diverse data. Despite recent efforts to develop foundational models for graph data~\cite{DBLP:conf/icml/MaoCT000S0T24}, these approaches remain limited to a few domains. Models~\cite{DBLP:conf/iclr/0057FKLT0Z24, DBLP:conf/iclr/0001YM0Z24} still fall short of providing a truly universal multi-modal graph encoding scheme, which hinders generalization to broader and more diverse application scenarios. Furthermore, the \textit{multi-granularity of nodes and structures} presents a significant challenge. In multi-modal graphs, nodes represent entities that may belong to different modalities (e.g., images, text, molecules) with various granularities, while edges capture relationships between these nodes. Effectively modeling these heterogeneous nodes and their connections, which often have varying structures, requires novel approaches for multi-modal embedding and multi-modal graph structure learning to better accommodate the complexity of diverse node types and relations. The highly irregular, multi-level, and potentially continuous nature of multi-graph vocabularies further complicates the creation of a unified representational space, leading to potential issues like \textit{redundancy} and \textit{information loss} during the integration of multi-modal features and relations.

\paragraph{Relevant literture}
Efforts towards achieving domain transferability and generalizable representations in graph learning have led to the development of foundation models and pre-training strategies for graph data~\cite{DBLP:conf/icml/MaoCT000S0T24, DBLP:conf/iclr/0057FKLT0Z24, DBLP:conf/iclr/0001YM0Z24}. Specifically for multi-modal graphs, research has begun exploring how to learn domain-invariant representations that can generalize across different knowledge graphs and multi-modal networks~\cite{zhu2025graphclip}. The growing interest in prompt-driven and instruction-based paradigms, largely influenced by the success of Large Language Models, has also spurred work in adapting these approaches for unified data processing within structured and multi-modal contexts~\cite{ye2024language, cao2025instructmol, liu2023visual}. While these works represent significant steps, they often grapple with unifying the vastly different granularities and semantic levels inherent in multi-modal graph data, or they rely on transformation functions that might incur information loss.

\paragraph{Future directions}
To achieve a truly unified space for multi-modal structures and attributes, several critical future directions emerge. Firstly, there is a pressing need to develop novel \textit{multi-modal graph vocabulary and tokenization schemes} that can flexibly represent information ranging from fine-grained features (e.g., pixels, words) to coarse-grained concepts (e.g., full images, entire documents), as well as complex graph topologies. This involves moving beyond simple concatenations or late fusion, aiming for an early and deep integration of multi-modal signals. Secondly, research should focus on designing architectures capable of learning genuinely \textit{transferable patterns} across the highly irregular, multi-level, and continuous nature of real-world multi-modal graphs. This might involve exploring advanced graph neural network designs combined with foundation models that can process diverse modalities natively. Thirdly, strategies to mitigate \textit{redundancy} and \textit{information loss} during multi-modal feature and relation integration are crucial. This could involve attention mechanisms that dynamically weigh modality contributions or latent space learning that preserves critical inter-modal dependencies. Finally, developing benchmark datasets specifically designed to evaluate the effectiveness of models in this unified multi-modal graph space, across diverse domains, will be essential to drive progress in this field. This foundational work will be instrumental in realizing the vision of \textit{native modeling} capable of operating directly on rich, multi-modal graph data without substantial information loss.

\subsection{Handling diverse multi-modal graph tasks}

\paragraph{Desired characteristics} A multi-modal graph large language model should be adept at handling a vast array of tasks, moving beyond traditional discriminative objectives to embrace a unified generative paradigm. 
As previously outlined, the ability to frame \textit{all multi-modal graph tasks as multi-modal graph generation} is a key characteristic. This necessitates a model capable of treating diverse problems, such as multi-modal node classification, link prediction, graph classification, question answering, reasoning, and even multi-modal content generation (text or image), as transformations from an input multi-modal graph to an output multi-modal graph. The model should demonstrate flexibility in its input and output spaces, seamlessly adapting to tasks that operate on different granularities, from fine-grained node features to entire graph structures. Furthermore, the model should support \textit{prompt-driven} or \textit{instruction-based learning}, allowing for versatile task adaptation and generalization to new, unseen multi-modal graph scenarios through natural language commands or structured prompts.

\paragraph{Key challenges}
A fundamental challenge in developing such models lies in the \textit{multi-scale characteristics} of multi-modal graph tasks. As highlighted in the generative modeling section, the input and output graphs can differ significantly in scope and granularity. For instance, a task might take a comprehensive multi-modal graph as input but require only a single predicted node’s attribute as output (as in node classification or a specific graph question answering task) \cite{yoon2023multimodal}. Conversely, other tasks might demand the generation of an entire sub-graph or even a new multi-modal graph from a simple input. This disparity in scales complicates \textit{unified task modeling} and especially \textit{task prompt design}, as the semantic levels of inputs and outputs vary widely \cite{sun2023all}. Beyond structural scale, the inherent scales of different modalities also pose challenges; text inputs can vary greatly in length, images in size, and videos in temporal duration. Integrating and generating information across these vastly different intrinsic modal scales, alongside varying graph structural complexities, requires sophisticated mechanisms to prevent information loss or redundancy and maintain coherence across the entire representation \cite{zhang2020multimodal}.

\paragraph{Relevant literture}
Efforts to address the diversity of graph tasks have led to the development of multi-task graph foundation models, which aim to provide a single framework for various graph-related tasks ~\cite{qiu2020gcc,sun2022gppt,liu2023graphprompt, yan2024inductive}. Inspired by the success of large language models, research has also begun to explore prompt-driven and instruction-based paradigms for graph and multi-modal contexts. These approaches leverage the flexibility of language models to adapt to different downstream tasks by formulating them as sequence-to-sequence or graph-to-sequence problems~\cite{yu2024multigprompt,zhao2024all}. Current GraphLLMs or vision-language models, by virtue of their flexible input and output capabilities, can handle some multi-modal graph tasks by first transforming the graph into a text-centric or image-text paired representation. While these methods demonstrate promising abilities in task generalization and leveraging pre-trained knowledge, they often face limitations when directly handling the intricate multi-scale nature and inherent graph structures, sometimes relying on transformation functions that may incur information loss or struggle with extremely long or complex contexts.

\paragraph{Future directions}
To effectively handle diverse multi-modal graph tasks, future research should prioritize developing novel approaches for \textit{unified task modeling} that natively account for the varying scales and granularities of both input and output multi-modal graphs. This involves designing architectures that can process fine-grained features (e.g., pixels, words) alongside coarse-grained concepts (e.g., full images, entire documents) and complex graph topologies in a seamless manner. Another critical direction is the investigation of \textit{adaptive prompting mechanisms} that can dynamically adjust to the task's specific scale and the required output granularity, moving beyond generic prompts. Furthermore, attention should be given to extending generative modeling techniques to truly \textit{open-set multi-modal graph generation}, where the model can synthesize novel graph structures or content (e.g., images, texts, audios) that are not limited to predefined sets. This requires a deeper integration of multi-modal understanding with generative capabilities, aiming for models that can operate directly on rich multi-modal graph information without relying on lossy intermediate transformations.

\subsection{Multi-modal graph in-context learning capability}
\paragraph{Desired characteristics}
A central aspiration for multi-modal graph large language models (\modelsnosp) is to exhibit robust \textit{in-context learning (ICL)} capabilities. This involves the model's ability to solve novel tasks by conditioning on a limited number of graph-anchored examples provided directly within the prompt, without requiring explicit weight updates or fine-tuning. Similar to how large language models learn from demonstrations in text, \model should infer underlying patterns and generalize to unseen multi-modal graph scenarios. This necessitates a model that can interpret and leverage diverse multi-modal features and complex graph structures present in the few-shot examples to inform its predictions or generations for new queries. Ultimately, achieving this capability relies on effective \textit{generative pretraining} on large-scale, paired multi-modal graph data, coupled with an architectural design and self-supervised learning objectives that facilitate flexible transfer and scaling.

\paragraph{Key challenges}
Extending in-context learning to graph-based multi-modal contexts presents significant challenges that are not encountered in plain text domains. Graphs inherently possess a \textit{variable topology}, \textit{long-range dependencies}, and a \textit{non-sequential structure}~\cite{bronstein2017geometric, peng2024learningmultimodalgraphssurvey}, making it difficult to define what constitutes an effective context window for ICL. Unlike a linear sequence of tokens, the `neighborhood' or `context' around a graph element can be complex and multifaceted. Furthermore, encoding the rich \textit{relational priors} and intricate inter-modal connections compactly for consumption by models, especially transformer architectures that are often designed for sequential data, remains a non-trivial task. The diversity of modalities within a graph (e.g., text, images, audio associated with nodes or edges) further complicates the unified representation and contextual understanding necessary for effective in-context learning.

\paragraph{Relevant literture}
Inspired by the success of large language models in in-context learning~\cite{brown2020language}, researchers have begun exploring methods to imbue this capability into graph-based models. Strategies for graph ICL often involve \textit{linearization strategies}, such as converting graphs or subgraphs into sequences of triples or using template filling to represent graph information in a text-like format~\cite{ye2024language}. Another approach involves the use of \textit{sampled subgraph prompts}, where relevant subgraphs are selected to serve as examples for the model~\cite{sun2022does}. \textit{Hybrid architectures} have also emerged, combining the strengths of pretrained graph neural networks (GNNs) for structural encoding with autoregressive decoders, which are adept at sequence generation and ICL~\cite{huang2024gnn}. Specific methods like AskGNN~\cite{sun2022does} and retrieval-augmented transformers adapted for graphs~\cite{lewis2020retrieval} demonstrate that carefully selecting relevant subgraphs and aligning them with language tokens can enhance few-shot learning within graph constraints, pointing towards the importance of context retrieval and integration. Progress has also been made in aligning graph information with textual prompts for language models~\cite{tang2024graphgpt, yu2025graph2text}.

\paragraph{Future directions}
To fully unlock the multi-modal graph in-context learning capability, several critical future directions should be explored. A key area is the development of more sophisticated \textit{multi-modal graph tokenization schemes} that can capture the inherent multi-granularity of graph entities (from fine-grained features to coarse-grained concepts) and their complex inter-modal relations in a way that is amenable to in-context learning. This would involve designing tokenizers that can flexibly abstract both structural and attribute information across modalities. Furthermore, research should focus on architectures that can intrinsically process irregular graph structures and multi-modal information without significant information loss from linearizing or simplifying the graph. This might involve novel graph-specific attention mechanisms or prompt-based techniques that can dynamically integrate and reason over graph-anchored examples. The integration of \textit{retrieval mechanisms} that can efficiently fetch relevant subgraphs or multi-modal contexts for ICL, especially from vast multi-modal knowledge graphs, will also be crucial. Ultimately, continued advancement in \textit{large-scale generative pretraining} on diverse and rich multi-modal graph datasets, coupled with self-supervised objectives tailored for graph understanding and generation, will be essential for models to acquire robust and transferable in-context learning abilities.


\subsection{Natural multi-modal graph interaction}
\paragraph{Desired characteristics}
An essential goal of Multi-Modal Graph Large Language Models (\modelsnosp) is to enable natural language-based interaction over structured and multi-modal data. Users should be able to query, edit, and reason about graph-structured knowledge using plain language, without needing to learn formal query languages like SPARQL or Cypher. This requires the \model to accurately map natural language inputs to graph traversal operations, complex reasoning chains, or precise node and edge modifications. Furthermore, the model should support rich, multi-turn dialogue, allowing for clarification and refinement of user intentions. A crucial aspect is \textit{visual grounding} within graphs, enabling the model to align natural language descriptions with visual elements present within the graph context, thereby facilitating a more intuitive understanding of multi-modal information. Beyond querying, the model should also be capable of graph-based summarization and generation of new graph structures or content in response to natural language commands. Ultimately, the desired characteristic is to provide a seamless, intuitive, and highly interactive interface between human users and complex multi-modal graph data, allowing for direct understanding, editing, reasoning, and generation of information, bridging the gap from unstructured human input to structured graph knowledge, while aligning with human values and intentions.

\paragraph{Key challenges}
One of the fundamental challenges in achieving natural multi-modal graph interaction lies in the inherent \textit{semantic gap} between the ambiguity and flexibility of natural language and the precise, structured nature of graph data. Natural language expressions can be vague or underspecified, making it difficult for a model to infer the user’s exact intention for graph operations, especially when dealing with heterogeneous multi-modal features~\cite{ektefaie2023multimodal}. Mapping these vague intentions to concrete graph traversals, edits, or reasoning chains is non-trivial. Furthermore, supporting multi-modal interactions introduces complexities, as the model must seamlessly interpret queries that might refer to text, image, audio, or video attributes within the graph, and potentially generate responses in a desired modality~\cite{lee2024multimodal}. The ability to \textit{understand, edit, reason, and generate} content within a multi-modal graph poses distinct challenges; for instance, editing an image-attributed node based on a textual command requires sophisticated cross-modal understanding and generative capabilities~\cite{chai2023graphllm}. Ensuring consistency and avoiding unintended side effects during graph modifications initiated by natural language commands is another significant hurdle. Finally, scaling such interactive capabilities to complex, real-world multi-modal graphs, such as those representing industry traffic patterns, molecular structures, or visual relational graphs, presents significant challenges in terms of computational efficiency and maintaining accuracy across diverse domains~\cite{li2025graph}.

\paragraph{Relevant literture}
Recent advancements in natural language interfaces for structured data have laid the foundational groundwork for multi-modal graph interaction. Efforts in \textit{conversational knowledge graphs}~\cite{xia2022lingyimedicalconversationalquestion, huang2023evaluating} have explored how to enable multi-turn dialogue over knowledge bases, allowing users to progressively refine their queries. Similarly, research on \textit{natural language interfaces for databases (NLIDB)}~\cite{liu2025nli4db} focuses on translating natural language questions into structured query languages, a precursor to graph traversal and modification. With the rise of multi-modal large language models, there has been increasing interest in \textit{visual grounding}, where models align language descriptions with visual elements in complex scenes or contexts~\cite{dong2024modality, huang2022endowinglanguagemodelsmultimodal, chen2022multi}. These techniques are directly relevant for grounding natural language queries within image or video attributed graph nodes. Graph-based generative tasks, such as \textit{graph summarization}~\cite{tang2024graphgpt, edge2025localglobalgraphrag}, have also emerged, demonstrating the ability to condense graph information into coherent text. Moreover, the integration of \textit{instruction-tuned language models with symbolic reasoning modules}~\cite{creswell2023selectioninference, wei2023symbol} represents a promising direction for building more robust dialogue-centric \modelsnosp, enabling them to leverage both pattern matching and logical deduction for complex graph interactions.

\paragraph{Future directions}
To fully realize natural multi-modal graph interaction, several critical future directions need exploration. Firstly, developing more sophisticated methods for \textit{disambiguating vague or underspecified natural language intentions} and aligning them precisely with multi-modal graph structures and attributes is crucial. This could involve interactive clarification dialogues where the model asks follow-up questions to refine its understanding. Secondly, research should focus on robust mechanisms for \textit{human feedback integration} during the interaction process, allowing users to correct model interpretations or outputs, thereby continually improving the model's understanding and alignment with human values. This iterative feedback loop is essential for adapting models to new domains and user preferences. Thirdly, advancing generative capabilities to enable not only querying but also \textit{complex graph editing and generation} through natural language commands is vital. This includes the ability to modify existing multi-modal nodes and edges, add new entities, or even generate entire subgraphs based on high-level instructions, with applications in areas like molecular design or urban planning. Finally, scaling these interaction paradigms to \textit{dynamic and evolving multi-modal graphs} that represent real-world phenomena (e.g., real-time industry traffic, evolving scientific knowledge graphs) will require innovations in efficient graph indexing, retrieval, and incremental updates to maintain responsiveness and accuracy during natural language-driven interactions.

\subsection{Multi-modal graph reasoning}

\paragraph{Desired characteristics}
An \model should be capable of \textit{multi-hop, cross-modal reasoning}. For example, the model should answer a complex query by combining clues from text and images through multiple inferential hops. Recent benchmarks like MultiModalQA~\cite{talmor2021multimodalqa} show that solving such cross-modal multi-hop questions remains challenging. Models must jointly reason over different modalities and knowledge sources to succeed on these tasks. Another desired capability is \textit{analogical inference across modalities}, where the model draws structural comparisons between, for instance, an image pair and a text pair. Analogical reasoning is a fundamental aspect of human cognition. Initial studies suggest large models have some analogical ability. However, most prior work on analogies is single-modal. Multi-modal analogical reasoning is still in its early stages. Early efforts on multimodal analogies over knowledge graphs (e.g., the MARS benchmark~\cite{zhang2023multimodal}) illustrate both the potential and the difficulty of this skill. For instance,  demonstrate that even advanced multimodal LLMs struggle with visual analogies unless special prompting or training is provided. This result underscores the importance of analogical inference as a future \model capability~\cite{guo2024can}.

\paragraph{Key challenges}
Building a \model poses several major challenges. First, \textit{modality alignment} is difficult. The model must align and fuse information from heterogeneous sources such as images, text, and graphs into a coherent representation. Without explicit alignment mechanisms, an image’s contents may not correctly map to textual concepts, impeding reasoning. Techniques like contrastive image-text pre-training (e.g., CLIP) are often used to partially address this problem by embedding modalities in a shared space~\cite{radford2021learning}. Recent \model approaches include dedicated alignment modules for vision and language. For example, MR-MKG \cite{lee2024multimodal} employs a cross-modal alignment module to optimize image and text correspondences within a multimodal knowledge graph. Second, \textit{factual consistency} remains a critical issue. Multimodal LLMs are prone to hallucination and may produce inconsistent answers that conflict with factual knowledge. This problem worsens when the model must recall external knowledge, such as from a graph that it was not pre-trained on. Recent work has highlighted these hallucination problems and proposed evaluation benchmarks (e.g., MHaluBench~\cite{chen2024unified}) and detection frameworks to reduce them. Indeed, MR-MKG was motivated by the observation that vanilla LLMs often fabricate details about images due to missing visual knowledge and injects a multimodal knowledge graph to ground the model in reality \cite{lee2024multimodal}. A third challenge is the fragility of current processing pipelines. Many models operate in stages or rely on external tools, and errors in early steps can cascade. \cite{liu2024towards} note that fixed sub-models in current systems make them unable to recover from intermediate mistakes. Improving robustness and feedback mechanisms is therefore an important challenge.

\paragraph{Relevant Literture}
Several initial approaches to \model have been proposed to address these challenges. A common strategy is to integrate a \textit{knowledge graph (KG)} or graph neural network into the multimodal pipeline to better handle structured, relational information. For example, MR-MKG~\cite{lee2024multimodal} augments a vision-language model with a multimodal knowledge graph that contains nodes and relations spanning text and images. It uses a relation-aware graph neural network to encode the MMKG and injects these representations into the LLM to improve reasoning. Another line of work focuses on graph construction and alignment between modalities. MAIL~\cite{dong2024modality} constructs a scene graph from image objects and a concept graph from external knowledge, aligning them via shared entities and fusing them through a pseudo-siamese graph neural network. This enables reasoning over a combined multimodal graph and has shown strong results in knowledge-intensive visual QA. Beyond QA, graph-enhanced multimodal models have been applied to other domains. For example, Choi\etal ~\cite{choi2024multimodal} proposes a model for healthcare that injects patient-specific graphs into an LLM and uses GNN-based message passing to align clinical text, lab results, and images. These methods highlight the benefit of structured reasoning but also reveal engineering complexity, as each task may require tailored graph construction and alignment strategies \cite{chen2022hybrid}.

\paragraph{Future directions}
Future research on multi-modal graph reasoning should tackle several ambitious goals. First, novel graph representation strategies tailored explicitly for multi-modal contexts could be developed, going beyond current embedding approaches to represent complex interactions between modalities more intuitively. Second, dynamically constructed multi-modal graphs that adapt in real-time to the context or queries presented to the model may enhance reasoning efficiency and accuracy. Additionally, exploring scalable inference techniques specifically designed for large and dense multi-modal graphs is essential to overcoming the context-length limitations of current models. Finally, there is a significant opportunity to advance explainability by designing methods that produce interpretable reasoning paths within multi-modal graph structures, enabling users to better understand and trust the model's decision-making process.

\subsection{Discussion on scalability and computational efficiency}

\paragraph{Computational efficiency}
\rev{Compared with unimodal LLMs, MG-LLMs face significantly higher computational demands due to the need to jointly encode complex graph structures and diverse multi-modal signals. Large-scale pretraining introduces massive parameter counts and memory usage, making naive extensions of existing LLM or GNN architectures impractical~\cite{Kaiser2017,Zaheer2020}. To improve efficiency, several strategies can be adopted: (1) Parameter sharing across modalities to reduce redundancy (as explored in unified multi-domain models~\cite{Kaiser2017}), (2) Modular architectures that enable different sub-modules to specialize on particular modalities or functions (e.g., separate expert components for each modality~\cite{Jaegle2021}), and (3) Sparse or sampled attention mechanisms to focus computation on the most relevant subgraphs and modalities~\cite{Zaheer2020,Tay2020}. For inference, pruning redundant tokens and layers (using efficient transformer techniques~\cite{Tay2020}), designing lightweight multi-modal graph tokenizers, and incorporating retrieval-augmented mechanisms~\cite{Lewis2020} are promising directions to lower latency without sacrificing accuracy.}

\paragraph{Scalability}
\rev{Real-world multi-modal graphs often involve millions of nodes and edges, coupled with highly heterogeneous modalities and relations. Scaling MG-LLMs to such large graphs requires both algorithmic and system-level innovations. On the algorithmic side, graph sampling methods (e.g., neighbor sampling~\cite{Hamilton2017} and subgraph sampling~\cite{Zeng2020}) and hierarchical modeling (e.g., differentiable graph pooling~\cite{Ying2018}) can make training and inference feasible on large datasets by reducing the effective graph size per batch. On the system side, distributed training pipelines and memory-efficient representations (e.g., sparse adjacency matrices or quantized features) are indispensable to handle billion-scale graphs. Frameworks like DistDGL~\cite{Zheng2021} demonstrate hybrid CPU-GPU training to scale GNNs to extremely large graphs. Furthermore, scalability must also account for modality and task diversity: MG-LLMs should adapt to varying input granularities and output structures while maintaining consistent generalization. Progressive scaling strategies and curriculum learning techniques~\cite{Bengio2009} may help stabilize training as the model gradually expands to increasingly large and diverse multi-modal graph corpora.}

\paragraph{Deployment strategies}
\rev{Even if an MG-LLM can be trained successfully, deployment in real-world scenarios poses additional constraints. Many applications (such as biomedical analysis or recommendation) impose strict latency requirements and often operate under hardware limitations. To make deployment feasible, model compression and distillation can reduce MG-LLMs into lighter domain-specific variants that retain core capabilities. For example, knowledge distillation transfers knowledge from a large teacher model to a smaller student model~\cite{Hinton2015}, and has been effective in compressing deep models without major performance loss. Quantization and pruning techniques can further improve inference speed and memory footprint (e.g., 8-bit quantization and weight pruning as in Deep Compression~\cite{Han2016}), enabling deployment on resource-limited edge devices. Domain-adapted MG-LLMs (for example, a variant specialized for molecular graphs in chemistry~\cite{Ying2021} or for urban spatial graphs in planning) are another strategy to balance generality and efficiency by focusing on a narrower range of modalities/tasks. Finally, hybrid deployment pipelines can be employed, where heavy multi-modal graph computations are offloaded to powerful servers (cloud) while lightweight modules run on clients (edge devices). Such split-computing approaches (akin to the Neurosurgeon framework for splitting DNN workloads~\cite{Kang2017}) achieve a practical compromise between responsiveness and accuracy, ensuring that MG-LLM-based solutions can meet real-time constraints in production environments.}

\section{Multi-modal graph datasets}
\label{sec:dataset}

Recent years have witnessed the emergence of multimodal graph learning datasets, which enrich traditional graph structures by incorporating image, text, video, audio, and multi-omics data. These datasets facilitate more challenging and realistic graph learning tasks by providing heterogeneous node and edge attributes. In this review, we categorize representative benchmarks by task type and summarize their scale, modalities and domains. statistics can be summarized in Table \ref{tab:dataset}. \rev{The datasets can also be grouped based on the origin of their domains, reflecting the types of real-world data they are derived from. Social network–related datasets originate from user interactions such as e-commerce activities, book recommendations, online articles, and urban information, which capture patterns of social behavior and digital connectivity. Knowledge graph datasets stem from structured repositories of knowledge, including multi-modal knowledge bases, artistic relationships, biomedical repositories, and recipe step data, emphasizing their foundation in curated domain-specific resources. Scene graph datasets, on the other hand, originate from visual scenes where objects and their relationships are explicitly annotated, making them distinct in their focus on spatial and semantic structure within images, and are mainly used for visual graph question answering tasks. Such a categorization is summarized in Table~\ref{tab:category}.}

\begin{table*}[ht]
\centering
\caption{\rev{Overview of tasks and their corresponding multi-modal graph datasets. \underline{NC} denotes Node Classification, \underline{LP} Link Prediction, \underline{GC} Graph Classification, \underline{GQA} Graph Question Answering, \underline{GR} Graph Reasoning, \underline{TG} Text Generation, and \underline{IG} Image Generation.}}
\label{tab:dataset}

\resizebox{\linewidth}{!}{
\begin{tabular}{@{}cllll@{}}
\toprule
\textbf{Task} & \textbf{Dataset} & \textbf{Modalities} & \textbf{Scale} & \textbf{Domain} \\
\midrule

\multirow{9}{*}{NC} 
 & ELE Fashion \cite{zhu2025mosaicmodalitiescomprehensivebenchmark}         & Text+Vision        & 98K nodes, 20K edges               & E-commerce products \\
 & Books NC \cite{zhu2025mosaicmodalitiescomprehensivebenchmark}           & Text+Vision         & 684K nodes, 7M edges               & Book recommendation \\
 & G2MF-Urban \cite{TAO2025104353}         & Text+Vision         & 100K nodes, 2M edges               & Urban planning \\
 & Pan-Cancer Atlas \cite{hoadley2018cell}  & Multi-omics         & 11K samples from 33 cancer types   & Biomedical repository \\
 & TCGA-BRCA \cite{cancer2012comprehensive}          & Multi-omics         & 1,084 breast tumor samples         & Biomedical repository \\
 & OMG-NAS Tencent \cite{Cai_Wang_Li_Zhang_Zhu_2024}    & Text+Vision         & 8K nodes, 60K edges                & Website articles \\
 & OMG-NAS Amazon \cite{Cai_Wang_Li_Zhang_Zhu_2024}     & Text+Vision         & 100K nodes, 300K edges             & E-commerce products \\
\cmidrule{1-5}

\multirow{6}{*}{LP} 
 & Books LP \cite{zhu2025mosaicmodalitiescomprehensivebenchmark}           & Text+Vision         & 64K nodes, 3437K edges             & Book recommendation \\
 & Sports CoPurchase \cite{zhu2025mosaicmodalitiescomprehensivebenchmark}  & Text+Vision         & 50K nodes, 25K edges               & E-commerce products \\
 & Cloth CoPurchase \cite{zhu2025mosaicmodalitiescomprehensivebenchmark}   & Text+Vision         & 126K nodes, 951K edges             & E-commerce products \\
 & HyperGCL-Ecomm \cite{saifuddin2025hypergclmultimodalgraphcontrastive}     & Text+Vision         & 1M edges, 500M edges               & E-commerce products \\
 & VTKG-I\&C \cite{lee-etal-2023-vista}         & Text+Vision         & 130 entities, 842 triples          & Multi-modal KGs \\
 & TIVA-KG \cite{10.1145/3581783.3612266}           & Text+Vision+Audio   & 50K entities, 200K triples         & Multi-modal KGs \\
\cmidrule{1-5}

\multirow{2}{*}{GC} 
 & OMG-NAS Recipe \cite{Cai_Wang_Li_Zhang_Zhu_2024}     & Text+Vision         & 20K nodes, 160K edges              & Food recipes \\
 & Large-RG \cite{ijcai2022p482}          & Text+Vision         & 500K nodes                         & Food recipes \\
\cmidrule{1-5}

\multirow{3}{*}{GQA} 
 & GQA \cite{8953451}               & Text+Vision         & 113K images, 23M questions         & Scene graphs \\
 & CLEVR \cite{Johnson_2017_CVPR}             & Text+Vision         & 100K images, 1M questions          & Scene graphs \\
 & SceneGraph-VQA \cite{alonso2025visionlanguagemodelsstrugglealign}    & Text+Vision         & 50K images                         & Scene graphs \\
\cmidrule{1-5}

\multirow{3}{*}{GR} 
 & MARS\&MarKG \cite{zhang2023multimodal}       & Text+Vision         & 34K triples, 13K questions         & Multi-modal KGs \\
 & FB-ING-TXT \cite{mousselly2018multimodal}        & Text+Vision         & 6K entities                        & Multi-modal KGs \\
 & WN9-ING-TXT \cite{mousselly2018multimodal}       & Text+Vision         & 12K entities                       & Multi-modal KGs \\
\cmidrule{1-5}

\multirow{3}{*}{TG} 
 & VRF \cite{shirai-etal-2022-visual}               & Text+Vision         & 200 recipes, 89 actions            & Food recipes \\
 & MS Recipe Corpus \cite{lin-etal-2020-recipe}  & Text+Vision         & 4K dishes, 150K recipes            & Food recipes \\
 & Richpedia \cite{10.1007/978-3-030-41407-8_9}         & Text+Vision         & 3M entities                        & Multi-modal KGs \\
\cmidrule{1-5}

\multirow{3}{*}{IG} 
 & ART500K \cite{fang2025graphgpt}               & Text+Vision         & 311K nodes, 643M edges            & Artwork relationships  \\
 & Amazon Coview \cite{jin2024instructg2i}  & Text+Vision         & 178K nodes, 3M edges           & E-commerce products \\
 & Goodreads \cite{jin2024instructg2i}     & Text+Vision         & 93K nodes, 637K edges          & Book recommendation \\
\bottomrule
\end{tabular}
}
\end{table*}

\begin{table*}[h]
\centering
\caption{ \rev{Datasets categorized by their origin, domain, and their examples.}}
\label{tab:category}
\resizebox{\linewidth}{!}{
\begin{tabular}{lll}
\toprule
\textbf{Origin} & \textbf{Domain} & \textbf{Datasets} \\
\midrule
\multirow{4}{*}{Social Network} 
  & E-commerce products & \begin{tabular}[c]{@{}l@{}} 
    ELE Fashion \cite{zhu2025mosaicmodalitiescomprehensivebenchmark}, OMG-NAS Amazon \cite{Cai_Wang_Li_Zhang_Zhu_2024}, Sports CoPurchase \cite{zhu2025mosaicmodalitiescomprehensivebenchmark}, Cloth CoPurchase \cite{zhu2025mosaicmodalitiescomprehensivebenchmark}, \\ 
    HyperGCL-Ecomm \cite{saifuddin2025hypergclmultimodalgraphcontrastive}, Amazon Coview \cite{jin2024instructg2i}
    \end{tabular} \\
  & Book recommendation & Books NC \cite{zhu2025mosaicmodalitiescomprehensivebenchmark}, Books LP \cite{zhu2025mosaicmodalitiescomprehensivebenchmark}, Goodreads \cite{jin2024instructg2i} \\
  & Website articles & OMG-NAS Tencent \cite{Cai_Wang_Li_Zhang_Zhu_2024} \\
  & Urban planning & G2MF-Urban \cite{TAO2025104353} \\
\midrule
\multirow{4}{*}{Knowledge Graph} 
  & Multi-modal KGs & \begin{tabular}[c]{@{}l@{}}  VTKG-I\&C \cite{lee-etal-2023-vista}, TIVA-KG \cite{10.1145/3581783.3612266}, MARS\&MarKG \cite{zhang2023multimodal}, FB-ING-TXT \cite{mousselly2018multimodal}, \\
  WN9-ING-TXT \cite{mousselly2018multimodal}, Richpedia \cite{10.1007/978-3-030-41407-8_9} \end{tabular} \\
  & Artwork relationships & ART500K \cite{fang2025graphgpt} \\
  & Biomedical repository & Pan-Cancer Atlas \cite{hoadley2018cell}, TCGA-BRCA \cite{cancer2012comprehensive} \\
  & Food recipes & OMG-NAS Recipe \cite{Cai_Wang_Li_Zhang_Zhu_2024}, Large-RG \cite{ijcai2022p482}, VRF \cite{shirai-etal-2022-visual}, MS Recipe Corpus \cite{lin-etal-2020-recipe} \\
\midrule
\multirow{1}{*}{Scene Graph} 
  & Scene graphs & GQA \cite{8953451}, CLEVR \cite{Johnson_2017_CVPR}, SceneGraph-VQA \cite{alonso2025visionlanguagemodelsstrugglealign} \\
\bottomrule
\end{tabular}}
\end{table*}

\subsection{Node classification}

Node classification benchmarks evaluate a model's ability to predict node labels when each node carries multimodal attributes. ELE Fashion \cite{zhu2025mosaicmodalitiescomprehensivebenchmark} is a medium-scale e-commerce product graph comprising approximately 97,800 product nodes, each being annotated with a product title and a high-resolution image. 
Books NC \cite{zhu2025mosaicmodalitiescomprehensivebenchmark} includes roughly 685,300 book nodes, each with cover images and descriptions, and is annotated for ten book categories. G2MF-Urban \cite{TAO2025104353} is an urban planning graph with about 100,000 street nodes and 2,000,000 edges, where nodes incorporate overhead imagery and Point of Interest (POI) text for functional zone classification. In the biomedical domain, the Pan-Cancer Atlas \cite{hoadley2018cell} comprises approximately 11,286 tumor samples across 33 cancer types, profiled by multiple omics assays (mRNA, miRNA, DNA methylation, proteomics, CNV), and is used for pan-cancer molecular subtype classification; each sample is modeled as a node whose features are the concatenated multimodal measurements, with edges encoding biological relations. Similarly, TCGA-BRCA \cite{cancer2012comprehensive} contains around 1,084 breast tumor samples assayed on six platforms (genomic, epigenomic, transcriptomic, proteomic) and supports breast cancer subtype classification, also modeling each sample as a node with concatenated multimodal measurements and edges encoding biological relations. 
Additionally, the OMG-NAS framework \cite{Cai_Wang_Li_Zhang_Zhu_2024} evaluates two real-world out-of-distribution benchmarks: the Tencent graph from WeChat official accounts, which includes 8,000 article nodes and 60,000 user-view edges, with each node carrying head images and titles; and the Amazon review graph, featuring 100,000 review nodes and 300,000 co-review edges, combining product images and textual feedback.

\subsection{Link prediction}

Link prediction datasets require inferring missing or future edges in graphs with multimodal node features. Books LP \cite{zhu2025mosaicmodalitiescomprehensivebenchmark} comprises approximately 636,500 book nodes, each with cover images and descriptions, used for link inference. The Sports CoPurchase and Cloth CoPurchase datasets \cite{zhu2025mosaicmodalitiescomprehensivebenchmark} are co-purchase graphs containing 50,250 and 125,839 product nodes respectively, with each node annotated with product titles and images. HyperGCL-Ecomm \cite{saifuddin2025hypergclmultimodalgraphcontrastive} is a hypergraph dataset with 1,000,000 product nodes possessing multimodal features and 500,000,000 behavioral edges. VTKG-I and VTKG-C \cite{lee-etal-2023-vista} present two commonsense knowledge graphs (KGs), each with approximately 130 entities and 842 triples, where entities are associated with images and text, designed for knowledge graph completion tasks. TIVA KG \cite{10.1145/3581783.3612266} is a quad-modal knowledge graph containing roughly 50,000 entities and 200,000 triples, which combines text, image, video, and audio modalities for completion tasks. 

\subsection{Graph classification}

Graph classification datasets learn holistic representations for entire graphs with heterogeneous node and edge attributes. In the context of the OMG-NAS framework \cite{Cai_Wang_Li_Zhang_Zhu_2024}, a Recipe graph is utilized, comprised of approximately 20,000 recipe nodes and 160,000 ingredient/instruction edges, where images are partitioned into 16×16 patch nodes and text into word nodes. Separately, the Large-RG dataset \cite{ijcai2022p482} models a culinary graph containing over 500,000 recipe nodes linked by ingredient edges, enriched with image and textual attributes.

\subsection{Visual graph QA}

Visual reasoning datasets convert images into scene graphs paired with compositional questions to assess structured inference. GQA \cite{8953451} contains 113,018 real-world images, 22.7 million questions, and scene graphs annotated with functional programs. CLEVR \cite{Johnson_2017_CVPR} is a synthetic visual question answering benchmark consisting of 100,000 rendered RGB scenes paired with approximately 853,000 automatically generated question–answer pairs. Each scene is accompanied by a detailed scene graph that encodes object attributes and spatial relations, and every question is mapped to a functional program specifying the multi-step reasoning required, with the dataset designed to minimize biases and provide exhaustive annotations. SceneGraph-VQA \cite{alonso2025visionlanguagemodelsstrugglealign} comprises 50,000 scene graphs with QA pairs, combining object-relationship hierarchies, image regions, and textual questions.

\subsection{Graph reasoning}
Multimodal graph reasoning datasets integrate heterogeneous information sources to support complex reasoning tasks. MARS and MarKG \cite{zhang2023multimodal} serve as benchmark datasets for multimodal analogical reasoning. MARS contains 10,685 training, 1,228 validation, and 1,415 test instances, where each task instance is a visual–textual analogical quadruple requiring the prediction of a missing entity. MarKG is a supporting knowledge graph for MARS, containing 11,292 entities and 192 relations, with entities enriched by 76,424 images along with textual and visual descriptions. Furthermore, two datasets, WN9-IMG-TXT and FB-IMG-TXT \cite{mousselly2018multimodal}, are widely adopted multimodal knowledge graph benchmarks. WN9-IMG-TXT contains 6555 entities, while FB-IMG-TXT contains 11757 entities. In both datasets, each entity is associated with three modalities: structural graph information, images, and text.

\subsection{Text generation}

These datasets evaluate alignment or generation of text conditioned on multimodal workflows. Visual Recipe Flow \cite{shirai-etal-2022-visual} annotates 200 recipes with before–and-after image pairs for each action and is grounded in a recipe-flow graph designed for stepwise text generation. The Microsoft Multimodal Aligned Recipe Corpus \cite{lin-etal-2020-recipe} contains approximately 150,000 text–video alignments across 4,262 dishes, structured with cross-modal graphs between recipe steps and corresponding video segments for description tasks. Richpedia \cite{10.1007/978-3-030-41407-8_9} models a multimodal knowledge base containing over 1,000,000 entities, where relations exist between KG textual entities and image entities, among image entities, or between image entities and values such as pixel information. This dataset is designed to support applications such as semantic search and text generation.

\subsection{Image generation}

Image generation utilizing multimodal graphs aims to synthesize visual content by leveraging the interconnected textual and visual information inherent in these complex network structures. For example, ART500K \cite{fang2025graphgpt} curates an artwork domain with 311,288 creations and 643 million interconnections reflecting shared artists or styles, each accompanied by textual and visual information. Amazon Coview \cite{jin2024instructg2i} charts an e-commerce product landscape where 178,890 items are linked by 3 million connections derived from concurrent Browse patterns, each item possessing textual and visual descriptions. Lastly, Goodreads \cite{jin2024instructg2i} forms a book recommendation network of 93,475 literary works interconnected by 637,210 relationships that highlight their comparability, with each book entry including textual and visual elements.

\subsection{Summary}
In summary, these datasets could be useful for evaluating multi-modal graph-learning methods across diverse tasks, scales, and fusion mechanisms, thereby laying a groundwork for \modelsnosp. Despite this progress, the current volume of multi-modal graph datasets remains significantly smaller than that used for pre-training large language models. This disparity highlights the urgent need for the community to develop more effective methods for data collection and utilization. Furthermore, many existing multi-modal graph tasks are discriminative, and future efforts might be devoted to more generative multi-modal graph tasks to advance the evaluation and design of \modelsnosp.

\section{Related works}

\subsection{Graph representation learning}

\rev{Graph neural networks (GNNs) such as the Graph Convolutional Network (GCN) \cite{kipf2017semi}, Graph Attention Network (GAT) \cite{velickovic2018graph}, and Graph Isomorphism Network (GIN) \cite{xu2019powerful} have become foundational for learning representations on graph-structured data. These models aggregate information over node neighborhoods to enable tasks like node classification and link prediction. However, recent surveys note several persistent challenges: scaling GNNs to massive graphs \cite{zhang2023gnnacc}, ensuring robustness to adversarial or noisy inputs \cite{dai2024trustworthy}, and integrating multimodal data into graph models \cite{ektefaie2023multimodal} remain open problems. To address graph-specific modeling in a more general framework, recent work has begun to incorporate large language models (LLMs) into graph learning. For example, LLaGA \cite{chen2024llaga} reformulates graphs as token sequences for an LLM, GOFA \cite{kong2024gofa} interleaves GNN layers within a pretrained LLM, and GraphRAG \cite{han2025graphrag} augments retrieval-augmented generation with knowledge graphs. These GraphLLM approaches explicitly leverage graph topology in their design. In contrast, general multimodal LLMs (e.g., GPT-4 \cite{openai2023gpt4}) are built for text and image inputs and do not directly encode graph structure. In this paper, we propose MG-LLM, a novel model that integrates graph structure with multi-modal information modeling to overcome existing limitations in unified representation of multi-modal structures and attributes, handling diverse multi-modal graph tasks, enabling in-context learning, and supporting multi-modal graph reasoning.}

\subsection{Large language models and multimodal large language models}
\rev{The field of artificial intelligence has been fundamentally reshaped by the advent of Large Language Models (LLMs) like the GPT series~\cite{brown2020language}, LLaMA~\cite{grattafiori2024llama}, and Qwen~\cite{yang2025qwen3}, built upon the transformer architecture~\cite{vaswani2017attention} and scaling principles~\cite{brown2020language}. This success has expanded the frontier to Multimodal Large Language Models (MLLMs) such as the seminal open-source LLaVA~\cite{liu2023visual}, GPT-4V~\cite{openai2023gpt4vsystemcard}, Qwen-VL~\cite{bai2025qwen25vl}, InternVL~\cite{wang2025internvl35}, and GLM-4.5V~\cite{vteam2025glm45v}. The trend has further pushed towards \textit{Omni-MLLMs} that handle arbitrary modalities~\cite{jiang2025from}, with prominent examples including the closed-source model GPT-4o~\cite{hurst2024gpt} and Gemini 2.5 Pro~\cite{comanici2025gemini} and open-source efforts like One-LLM~\cite{han2024onellm} and CoDi-2~\cite{tang2024codi}. A parallel development is the rise of LLMs with agentic capabilities for tool use and planning, exemplified by models like Kimi K2~\cite{team2025kimi} and GLM-4.5~\cite{zeng2025glm}. Despite these significant advancements, a common limitation persists: current models are primarily designed for sequential data and lack a native mechanism for reasoning over explicit, complex relational structures. This highlights a critical gap in handling multi-modal graph topology, which our proposed MG-LLM aims to address.}

\section{Conclusion}
In this paper, we aim to address the generalization limits of current Multi-modal Graph Neural Networks by proposing Multi-modal Graph Large Language Models (\modelsnosp). We introduce a unified framework, highlighting the inherent characteristics of multi-modal graphs, and define five essential capabilities for \modelsnosp, ranging from unified data representation to complex reasoning. By discussing challenges, related research, and future directions, this work aims to contribute to the multi-modal graph community and accelerate the development of versatile, general-purpose \modelsnosp.

\Acknowledgements{This work was supported by the National Key Research and Development Program of China No.2023YFF1205001, National Natural Science Foundation of China (No. 62222209), Beijing National Research Center for Information Science and Technology under Grant No. BNR2023TD03006, and Beijing Key Lab of Networked Multimedia.}

\bibliographystyle{scis}

\begin{thebibliography}{100}
\providecommand{\url}[1]{\texttt{#1}}
\providecommand{\urlprefix}{URL }
\providecommand{\bibinfo}[2]{#2}

\bibitem{bhattacharyya2024heterogeneous}
\bibinfo{author}{Bhattacharyya S}, \bibinfo{author}{Yang S}, \bibinfo{author}{Wang J~Z}.
\newblock \bibinfo{title}{A heterogeneous multimodal graph learning framework for recognizing user emotions in social networks}.
\newblock In: \bibinfo{booktitle}{2024 12th International Conference on Affective Computing and Intelligent Interaction (ACII)}. \bibinfo{organization}{IEEE}\bibinfo{year}{2024}.
\newblock \bibinfo{pages}{327--336}

\bibitem{wang2023fashionklip}
\bibinfo{author}{Wang X}, \bibinfo{author}{Wang C}, \bibinfo{author}{Li L}, et~al.
\newblock \bibinfo{title}{Fashionklip: enhancing e-commerce image-text retrieval with fashion multi-modal conceptual knowledge graph}.
\newblock In: \bibinfo{booktitle}{Proceedings of the 61st Annual Meeting of the Association for Computational Linguistics (Volume 5: Industry Track)}, \bibinfo{year}{2023}.
\newblock \bibinfo{pages}{149--158}

\bibitem{marini2024multimodal}
\bibinfo{author}{Marini N}, \bibinfo{author}{Marchesin S}, \bibinfo{author}{Wodzinski M}, et~al.
\newblock \bibinfo{title}{Multimodal representations of biomedical knowledge from limited training whole slide images and reports using deep learning}.
\newblock \bibinfo{journal}{Medical Image Analysis}, \bibinfo{year}{2024}, \bibinfo{volume}{97}: \bibinfo{pages}{103303}

\bibitem{ye2024construction}
\bibinfo{author}{Ye Y}, \bibinfo{author}{Ren J}, \bibinfo{author}{Wang S}, et~al.
\newblock \bibinfo{title}{Construction and application of materials knowledge graph in multidisciplinary materials science via large language model}.
\newblock \bibinfo{journal}{Advances in Neural Information Processing Systems}, \bibinfo{year}{2024}, \bibinfo{volume}{37}: \bibinfo{pages}{56878--56897}

\bibitem{ektefaie2023multimodal}
\bibinfo{author}{Ektefaie Y}, \bibinfo{author}{Dasoulas G}, \bibinfo{author}{Noori A}, et~al.
\newblock \bibinfo{title}{Multimodal learning with graphs}.
\newblock \bibinfo{journal}{Nature Machine Intelligence}, \bibinfo{year}{2023}, \bibinfo{volume}{5}: \bibinfo{pages}{340--350}

\bibitem{peng2024learningmultimodalgraphssurvey}
\bibinfo{author}{Peng C}, \bibinfo{author}{He J}, \bibinfo{author}{Xia F}.
\newblock \bibinfo{title}{Learning on multimodal graphs: A survey}, \bibinfo{year}{2024}

\bibitem{chen2024llaga}
\bibinfo{author}{Chen R}, \bibinfo{author}{Zhao T}, \bibinfo{author}{Jaiswal A}, et~al.
\newblock \bibinfo{title}{Llaga: Large language and graph assistant}.
\newblock \bibinfo{journal}{arXiv preprint arXiv:2402.08170}, \bibinfo{year}{2024}: \bibinfo{pages}{7809–7823}

\bibitem{kong2024gofa}
\bibinfo{author}{Kong L}, \bibinfo{author}{Feng J}, \bibinfo{author}{Liu H}, et~al.
\newblock \bibinfo{title}{Gofa: A generative one-for-all model for joint graph language modeling}.
\newblock \bibinfo{journal}{arXiv preprint arXiv:2407.09709}, \bibinfo{year}{2024}

\bibitem{jiang2025from}
\bibinfo{author}{Jiang S}, \bibinfo{author}{Liang J}, \bibinfo{author}{Wang J}, et~al.
\newblock \bibinfo{title}{From specific-mllms to omni-mllms: A survey on mllms aligned with multi-modalities}.
\newblock In: \bibinfo{booktitle}{Findings of the Association for Computational Linguistics: ACL 2025}. \bibinfo{publisher}{Association for Computational Linguistics}\bibinfo{year}{2025}.
\newblock \bibinfo{pages}{8617--8652}

\bibitem{zhu2025mosaicmodalitiescomprehensivebenchmark}
\bibinfo{author}{Zhu J}, \bibinfo{author}{Zhou Y}, \bibinfo{author}{Qian S}, et~al.
\newblock \bibinfo{title}{Mosaic of modalities: A comprehensive benchmark for multimodal graph learning}, \bibinfo{year}{2025}

\bibitem{10.1007/978-3-030-41407-8_9}
\bibinfo{author}{Wang M}, \bibinfo{author}{Qi G}, \bibinfo{author}{Wang H}, et~al.
\newblock \bibinfo{title}{Richpedia: A comprehensive multi-modal knowledge graph}.
\newblock In: \bibinfo{booktitle}{Semantic Technology}. \bibinfo{publisher}{Springer International Publishing}\bibinfo{year}{2020}.
\newblock \bibinfo{pages}{130--145}

\bibitem{8953451}
\bibinfo{author}{Hudson D~A}, \bibinfo{author}{Manning C~D}.
\newblock \bibinfo{title}{Gqa: A new dataset for real-world visual reasoning and compositional question answering}.
\newblock In: \bibinfo{booktitle}{2019 IEEE/CVF Conference on Computer Vision and Pattern Recognition (CVPR)}, \bibinfo{year}{2019}.
\newblock \bibinfo{pages}{6693--6702}

\bibitem{zhao2024deep}
\bibinfo{author}{Zhao F}, \bibinfo{author}{Zhang C}, \bibinfo{author}{Geng B}.
\newblock \bibinfo{title}{Deep multimodal data fusion}.
\newblock \bibinfo{journal}{ACM computing surveys}, \bibinfo{year}{2024}, \bibinfo{volume}{56}: \bibinfo{pages}{1--36}

\bibitem{Wei2019MMGCN}
\bibinfo{author}{Wei Y}, \bibinfo{author}{Wang X}, \bibinfo{author}{Nie L}, et~al.
\newblock \bibinfo{title}{Mmgcn: Multi-modal graph convolution network for personalized recommendation of micro-video}.
\newblock In: \bibinfo{booktitle}{Proceedings of the 27th ACM International Conference on Multimedia (MM 2019)}. \bibinfo{publisher}{ACM}\bibinfo{year}{2019}.
\newblock \bibinfo{pages}{1437--1445}

\bibitem{Song2023Multicenter}
\bibinfo{author}{Song X}, \bibinfo{author}{Zhou F}, \bibinfo{author}{Frangi A~F}, et~al.
\newblock \bibinfo{title}{Multicenter and multichannel pooling gcn for early ad diagnosis based on dual-modality fused brain network}.
\newblock \bibinfo{journal}{IEEE Transactions on Medical Imaging}, \bibinfo{year}{2023}, \bibinfo{volume}{42}: \bibinfo{pages}{354--367}

\bibitem{Zeng2023Hypergraph}
\bibinfo{author}{Zeng Y}, \bibinfo{author}{Jin Q}, \bibinfo{author}{Bao T}, et~al.
\newblock \bibinfo{title}{Multi-modal knowledge hypergraph for diverse image retrieval}.
\newblock In: \bibinfo{booktitle}{Proceedings of the AAAI Conference on Artificial Intelligence}, \bibinfo{year}{2023}, volume~\bibinfo{volume}{37}.
\newblock \bibinfo{pages}{3376--3383}

\bibitem{tian2022recipe2vecmultimodalreciperepresentation}
\bibinfo{author}{Tian Y}, \bibinfo{author}{Zhang C}, \bibinfo{author}{Guo Z}, et~al.
\newblock \bibinfo{title}{Recipe2vec: Multi-modal recipe representation learning with graph neural networks}, \bibinfo{year}{2022}

\bibitem{he-wang-2023-multimodal}
\bibinfo{author}{He X}, \bibinfo{author}{Wang X}.
\newblock \bibinfo{title}{Multimodal graph transformer for multimodal question answering}.
\newblock In: \bibinfo{booktitle}{Proceedings of the 17th Conference of the European Chapter of the Association for Computational Linguistics}. \bibinfo{publisher}{Association for Computational Linguistics}\bibinfo{year}{2023}.
\newblock \bibinfo{pages}{189--200}

\bibitem{9950299}
\bibinfo{author}{Cai H}, \bibinfo{author}{Gao Y}, \bibinfo{author}{Liu M}.
\newblock \bibinfo{title}{Graph transformer geometric learning of brain networks using multimodal mr images for brain age estimation}.
\newblock \bibinfo{journal}{IEEE Transactions on Medical Imaging}, \bibinfo{year}{2023}, \bibinfo{volume}{42}: \bibinfo{pages}{456--466}

\bibitem{wang2024can}
\bibinfo{author}{Wang H}, \bibinfo{author}{Feng S}, \bibinfo{author}{He T}, et~al.
\newblock \bibinfo{title}{Can language models solve graph problems in natural language?}
\newblock \bibinfo{journal}{Advances in Neural Information Processing Systems}, \bibinfo{year}{2024}, \bibinfo{volume}{36}

\bibitem{guo2023gpt4graph}
\bibinfo{author}{Guo J}, \bibinfo{author}{Du L}, \bibinfo{author}{Liu H}, et~al.
\newblock \bibinfo{title}{Gpt4graph: Can large language models understand graph structured data? an empirical evaluation and benchmarking}.
\newblock \bibinfo{journal}{arXiv preprint arXiv:2305.15066}, \bibinfo{year}{2023}

\bibitem{zhao2023gimlet}
\bibinfo{author}{Zhao H}, \bibinfo{author}{Liu S}, \bibinfo{author}{Chang M}, et~al.
\newblock \bibinfo{title}{Gimlet: A unified graph-text model for instruction-based molecule zero-shot learning}.
\newblock \bibinfo{journal}{Advances in Neural Information Processing Systems}, \bibinfo{year}{2023}, \bibinfo{volume}{36}: \bibinfo{pages}{5850--5887}

\bibitem{liu2024git}
\bibinfo{author}{Liu P}, \bibinfo{author}{Ren Y}, \bibinfo{author}{Tao J}, et~al.
\newblock \bibinfo{title}{Git-mol: A multi-modal large language model for molecular science with graph, image, and text}.
\newblock \bibinfo{journal}{Computers in biology and medicine}, \bibinfo{year}{2024}, \bibinfo{volume}{171}: \bibinfo{pages}{108073}

\bibitem{liu2023evaluating}
\bibinfo{author}{Liu C}, \bibinfo{author}{Wu B}.
\newblock \bibinfo{title}{Evaluating large language models on graphs: Performance insights and comparative analysis}.
\newblock \bibinfo{journal}{arXiv preprint arXiv:2308.11224}, \bibinfo{year}{2023}

\bibitem{fatemi2023talk}
\bibinfo{author}{Fatemi B}, \bibinfo{author}{Halcrow J}, \bibinfo{author}{Perozzi B}.
\newblock \bibinfo{title}{Talk like a graph: Encoding graphs for large language models}.
\newblock \bibinfo{journal}{arXiv preprint arXiv:2310.04560}, \bibinfo{year}{2023}

\bibitem{hu2023beyond}
\bibinfo{author}{Hu Y}, \bibinfo{author}{Zhang Z}, \bibinfo{author}{Zhao L}.
\newblock \bibinfo{title}{Beyond text: A deep dive into large language models' ability on understanding graph data}.
\newblock \bibinfo{journal}{arXiv preprint arXiv:2310.04944}, \bibinfo{year}{2023}

\bibitem{cai2024multimodalgraph}
\bibinfo{author}{Cai J}, \bibinfo{author}{Wang X}, \bibinfo{author}{Li H}, et~al.
\newblock \bibinfo{title}{Multimodal graph neural architecture search under distribution shifts}.
\newblock \bibinfo{journal}{Proceedings of the AAAI Conference on Artificial Intelligence}, \bibinfo{year}{2024}, \bibinfo{volume}{38}: \bibinfo{pages}{8227--8235}

\bibitem{zitnik2018modeling}
\bibinfo{author}{Zitnik M}, \bibinfo{author}{Agrawal M}, \bibinfo{author}{Leskovec J}.
\newblock \bibinfo{title}{Modeling polypharmacy side effects with graph convolutional networks}.
\newblock \bibinfo{journal}{Bioinformatics}, \bibinfo{year}{2018}, \bibinfo{volume}{34}: \bibinfo{pages}{i457--i466}

\bibitem{wu2020comprehensive}
\bibinfo{author}{Wu Z}, \bibinfo{author}{Pan S}, \bibinfo{author}{Chen F}, et~al.
\newblock \bibinfo{title}{A comprehensive survey on graph neural networks}.
\newblock \bibinfo{journal}{IEEE transactions on neural networks and learning systems}, \bibinfo{year}{2020}, \bibinfo{volume}{32}: \bibinfo{pages}{4--24}

\bibitem{wang2021kepler}
\bibinfo{author}{Wang X}, \bibinfo{author}{Gao T}, \bibinfo{author}{Zhu Z}, et~al.
\newblock \bibinfo{title}{Kepler: A unified model for knowledge embedding and pre-trained language representation}.
\newblock \bibinfo{journal}{Transactions of the Association for Computational Linguistics}, \bibinfo{year}{2021}, \bibinfo{volume}{9}: \bibinfo{pages}{176--194}

\bibitem{srivastava2014dropout}
\bibinfo{author}{Srivastava N}, \bibinfo{author}{Hinton G}, \bibinfo{author}{Krizhevsky A}, et~al.
\newblock \bibinfo{title}{Dropout: a simple way to prevent neural networks from overfitting}.
\newblock \bibinfo{journal}{The journal of machine learning research}, \bibinfo{year}{2014}, \bibinfo{volume}{15}: \bibinfo{pages}{1929--1958}

\bibitem{DBLP:conf/icml/MaoCT000S0T24}
\bibinfo{author}{Mao H}, \bibinfo{author}{Chen Z}, \bibinfo{author}{Tang W}, et~al.
\newblock \bibinfo{title}{Position: Graph foundation models are already here}.
\newblock In: \bibinfo{booktitle}{Forty-first International Conference on Machine Learning, {ICML} 2024, Vienna, Austria, July 21-27, 2024}. \bibinfo{publisher}{OpenReview.net}\bibinfo{year}{2024}

\bibitem{DBLP:conf/iclr/0057FKLT0Z24}
\bibinfo{author}{Liu H}, \bibinfo{author}{Feng J}, \bibinfo{author}{Kong L}, et~al.
\newblock \bibinfo{title}{One for all: Towards training one graph model for all classification tasks}.
\newblock In: \bibinfo{booktitle}{The Twelfth International Conference on Learning Representations, {ICLR} 2024, Vienna, Austria, May 7-11, 2024}. \bibinfo{publisher}{OpenReview.net}\bibinfo{year}{2024}

\bibitem{DBLP:conf/iclr/0001YM0Z24}
\bibinfo{author}{Galkin M}, \bibinfo{author}{Yuan X}, \bibinfo{author}{Mostafa H}, et~al.
\newblock \bibinfo{title}{Towards foundation models for knowledge graph reasoning}.
\newblock In: \bibinfo{booktitle}{The Twelfth International Conference on Learning Representations, {ICLR} 2024, Vienna, Austria, May 7-11, 2024}. \bibinfo{publisher}{OpenReview.net}\bibinfo{year}{2024}

\bibitem{zhu2025graphclip}
\bibinfo{author}{Zhu Y}, \bibinfo{author}{Shi H}, \bibinfo{author}{Wang X}, et~al.
\newblock \bibinfo{title}{Graphclip: Enhancing transferability in graph foundation models for text-attributed graphs}, \bibinfo{year}{2025}

\bibitem{ye2024language}
\bibinfo{author}{Ye R}, \bibinfo{author}{Zhang C}, \bibinfo{author}{Wang R}, et~al.
\newblock \bibinfo{title}{Language is all a graph needs}.
\newblock In: \bibinfo{booktitle}{Findings of the Association for Computational Linguistics: EACL 2024}. \bibinfo{publisher}{Association for Computational Linguistics}\bibinfo{year}{2024}.
\newblock \bibinfo{pages}{1955--1973}

\bibitem{cao2025instructmol}
\bibinfo{author}{Cao H}, \bibinfo{author}{Liu Z}, \bibinfo{author}{Lu X}, et~al.
\newblock \bibinfo{title}{Instructmol: Multi-modal integration for building a versatile and reliable molecular assistant in drug discovery}.
\newblock In: \bibinfo{booktitle}{Proceedings of the 31st International Conference on Computational Linguistics}. \bibinfo{publisher}{Association for Computational Linguistics}\bibinfo{year}{2025}.
\newblock \bibinfo{pages}{354--379}

\bibitem{liu2023visual}
\bibinfo{author}{Liu H}, \bibinfo{author}{Li C}, \bibinfo{author}{Wu Q}, et~al.
\newblock \bibinfo{title}{Visual instruction tuning}.
\newblock \bibinfo{journal}{Advances in neural information processing systems}, \bibinfo{year}{2023}, \bibinfo{volume}{36}: \bibinfo{pages}{34892--34916}

\bibitem{yoon2023multimodal}
\bibinfo{author}{Yoon M}, \bibinfo{author}{Koh J~Y}, \bibinfo{author}{Hooi B}, et~al.
\newblock \bibinfo{title}{Multimodal graph learning for generative tasks}.
\newblock \bibinfo{journal}{arXiv preprint arXiv:2310.07478}, \bibinfo{year}{2023}

\bibitem{sun2023all}
\bibinfo{author}{Sun X}, \bibinfo{author}{Cheng H}, \bibinfo{author}{Li J}, et~al.
\newblock \bibinfo{title}{All in one: Multi-task prompting for graph neural networks}.
\newblock In: \bibinfo{booktitle}{Proceedings of the 29th ACM SIGKDD Conference on Knowledge Discovery and Data Mining}, \bibinfo{year}{2023}.
\newblock \bibinfo{pages}{2120--2131}

\bibitem{zhang2020multimodal}
\bibinfo{author}{Zhang C}, \bibinfo{author}{Yang Z}, \bibinfo{author}{He X}, et~al.
\newblock \bibinfo{title}{Multimodal intelligence: Representation learning, information fusion, and applications}.
\newblock \bibinfo{journal}{IEEE Journal of Selected Topics in Signal Processing}, \bibinfo{year}{2020}, \bibinfo{volume}{14}: \bibinfo{pages}{478--493}

\bibitem{qiu2020gcc}
\bibinfo{author}{Qiu J}, \bibinfo{author}{Chen Q}, \bibinfo{author}{Dong Y}, et~al.
\newblock \bibinfo{title}{Gcc: Graph contrastive coding for graph neural network pre-training}.
\newblock In: \bibinfo{booktitle}{Proceedings of the 26th ACM SIGKDD international conference on knowledge discovery \& data mining}, \bibinfo{year}{2020}.
\newblock \bibinfo{pages}{1150--1160}

\bibitem{sun2022gppt}
\bibinfo{author}{Sun M}, \bibinfo{author}{Zhou K}, \bibinfo{author}{He X}, et~al.
\newblock \bibinfo{title}{Gppt: Graph pre-training and prompt tuning to generalize graph neural networks}.
\newblock In: \bibinfo{booktitle}{Proceedings of the 28th ACM SIGKDD Conference on Knowledge Discovery and Data Mining}, \bibinfo{year}{2022}.
\newblock \bibinfo{pages}{1717--1727}

\bibitem{liu2023graphprompt}
\bibinfo{author}{Liu Z}, \bibinfo{author}{Yu X}, \bibinfo{author}{Fang Y}, et~al.
\newblock \bibinfo{title}{Graphprompt: Unifying pre-training and downstream tasks for graph neural networks}.
\newblock In: \bibinfo{booktitle}{Proceedings of the ACM Web Conference 2023}, \bibinfo{year}{2023}.
\newblock \bibinfo{pages}{417--428}

\bibitem{yan2024inductive}
\bibinfo{author}{Yan Y}, \bibinfo{author}{Zhang P}, \bibinfo{author}{Fang Z}, et~al.
\newblock \bibinfo{title}{Inductive graph alignment prompt: Bridging the gap between graph pre-training and inductive fine-tuning from spectral perspective}.
\newblock In: \bibinfo{booktitle}{Proceedings of the ACM on Web Conference 2024}, \bibinfo{year}{2024}.
\newblock \bibinfo{pages}{4328--4339}

\bibitem{yu2024multigprompt}
\bibinfo{author}{Yu X}, \bibinfo{author}{Zhou C}, \bibinfo{author}{Fang Y}, et~al.
\newblock \bibinfo{title}{Multigprompt for multi-task pre-training and prompting on graphs}.
\newblock In: \bibinfo{booktitle}{Proceedings of the ACM on Web Conference 2024}, \bibinfo{year}{2024}.
\newblock \bibinfo{pages}{515--526}

\bibitem{zhao2024all}
\bibinfo{author}{Zhao H}, \bibinfo{author}{Chen A}, \bibinfo{author}{Sun X}, et~al.
\newblock \bibinfo{title}{All in one and one for all: A simple yet effective method towards cross-domain graph pretraining}.
\newblock In: \bibinfo{booktitle}{Proceedings of the 30th ACM SIGKDD Conference on Knowledge Discovery and Data Mining}, \bibinfo{year}{2024}.
\newblock \bibinfo{pages}{4443--4454}

\bibitem{bronstein2017geometric}
\bibinfo{author}{Bronstein M~M}, \bibinfo{author}{Bruna J}, \bibinfo{author}{LeCun Y}, et~al.
\newblock \bibinfo{title}{Geometric deep learning: going beyond euclidean data}.
\newblock \bibinfo{journal}{IEEE Signal Processing Magazine}, \bibinfo{year}{2017}, \bibinfo{volume}{34}: \bibinfo{pages}{18--42}

\bibitem{brown2020language}
\bibinfo{author}{Brown T}, \bibinfo{author}{Mann B}, \bibinfo{author}{Ryder N}, et~al.
\newblock \bibinfo{title}{Language models are few-shot learners}.
\newblock \bibinfo{journal}{Advances in neural information processing systems}, \bibinfo{year}{2020}, \bibinfo{volume}{33}: \bibinfo{pages}{1877--1901}

\bibitem{sun2022does}
\bibinfo{author}{Sun R}, \bibinfo{author}{Dai H}, \bibinfo{author}{Yu A~W}.
\newblock \bibinfo{title}{Does gnn pretraining help molecular representation?}
\newblock \bibinfo{journal}{Advances in Neural Information Processing Systems}, \bibinfo{year}{2022}, \bibinfo{volume}{35}: \bibinfo{pages}{12096--12109}

\bibitem{huang2024gnn}
\bibinfo{author}{Huang X}, \bibinfo{author}{Han K}, \bibinfo{author}{Yang Y}, et~al.
\newblock \bibinfo{title}{Can gnn be good adapter for llms?}
\newblock In: \bibinfo{booktitle}{Proceedings of the ACM Web Conference 2024}. \bibinfo{publisher}{Association for Computing Machinery}\bibinfo{year}{2024}, WWW '24.
\newblock \bibinfo{pages}{893–904}

\bibitem{lewis2020retrieval}
\bibinfo{author}{Lewis P}, \bibinfo{author}{Perez E}, \bibinfo{author}{Piktus A}, et~al.
\newblock \bibinfo{title}{Retrieval-augmented generation for knowledge-intensive nlp tasks}.
\newblock \bibinfo{journal}{Advances in neural information processing systems}, \bibinfo{year}{2020}, \bibinfo{volume}{33}: \bibinfo{pages}{9459--9474}

\bibitem{tang2024graphgpt}
\bibinfo{author}{Tang J}, \bibinfo{author}{Yang Y}, \bibinfo{author}{Wei W}, et~al.
\newblock \bibinfo{title}{Graphgpt: Graph instruction tuning for large language models}.
\newblock In: \bibinfo{booktitle}{Proceedings of the 47th International ACM SIGIR Conference on Research and Development in Information Retrieval}, \bibinfo{year}{2024}.
\newblock \bibinfo{pages}{491--500}

\bibitem{yu2025graph2text}
\bibinfo{author}{Yu S}, \bibinfo{author}{Wang Y}, \bibinfo{author}{Li R}, et~al.
\newblock \bibinfo{title}{Graph2text or graph2token: A perspective of large language models for graph learning}.
\newblock \bibinfo{journal}{arXiv preprint arXiv:2501.01124}, \bibinfo{year}{2025}

\bibitem{lee2024multimodal}
\bibinfo{author}{Lee J}, \bibinfo{author}{Wang Y}, \bibinfo{author}{Li J}, et~al.
\newblock \bibinfo{title}{Multimodal reasoning with multimodal knowledge graph}.
\newblock In: \bibinfo{booktitle}{Proceedings of the 62nd Annual Meeting of the Association for Computational Linguistics (Volume 1: Long Papers)}. \bibinfo{publisher}{Association for Computational Linguistics}\bibinfo{year}{2024}.
\newblock \bibinfo{pages}{10767--10782}

\bibitem{chai2023graphllm}
\bibinfo{author}{Chai Z}, \bibinfo{author}{Zhang T}, \bibinfo{author}{Wu L}, et~al.
\newblock \bibinfo{title}{Graphllm: Boosting graph reasoning ability of large language model}.
\newblock \bibinfo{journal}{arXiv preprint arXiv:2310.05845}, \bibinfo{year}{2023}

\bibitem{li2025graph}
\bibinfo{author}{Li R}, \bibinfo{author}{Jiang H}.
\newblock \bibinfo{title}{Graph-to-vision: Multi-graph understanding and reasoning using vision-language models}.
\newblock \bibinfo{journal}{arXiv preprint arXiv:2503.21435}, \bibinfo{year}{2025}

\bibitem{xia2022lingyimedicalconversationalquestion}
\bibinfo{author}{Xia F}, \bibinfo{author}{Li B}, \bibinfo{author}{Weng Y}, et~al.
\newblock \bibinfo{title}{Lingyi: medical conversational question answering system based on multi-modal knowledge graphs}.
\newblock \bibinfo{journal}{arXiv preprint arXiv:2204.09220}, \bibinfo{year}{2022}

\bibitem{huang2023evaluating}
\bibinfo{author}{Huang Y}, \bibinfo{author}{Shi L}, \bibinfo{author}{Liu A}, et~al.
\newblock \bibinfo{title}{Evaluating and enhancing large language models for conversational reasoning on knowledge graphs}.
\newblock \bibinfo{journal}{arXiv preprint arXiv:2312.11282}, \bibinfo{year}{2023}

\bibitem{liu2025nli4db}
\bibinfo{author}{Liu M}, \bibinfo{author}{Xu J}.
\newblock \bibinfo{title}{Nli4db: A systematic review of natural language interfaces for databases}, \bibinfo{year}{2025}

\bibitem{dong2024modality}
\bibinfo{author}{Dong J}, \bibinfo{author}{Zhang Q}, \bibinfo{author}{Zhou H}, et~al.
\newblock \bibinfo{title}{Modality-aware integration with large language models for knowledge-based visual question answering}.
\newblock In: \bibinfo{booktitle}{Proceedings of the 62nd Annual Meeting of the Association for Computational Linguistics (Volume 1: Long Papers)}. \bibinfo{publisher}{Association for Computational Linguistics}\bibinfo{year}{2024}.
\newblock \bibinfo{pages}{2417--2429}

\bibitem{huang2022endowinglanguagemodelsmultimodal}
\bibinfo{author}{Huang N}, \bibinfo{author}{Deshpande Y~R}, \bibinfo{author}{Liu Y}, et~al.
\newblock \bibinfo{title}{Endowing language models with multimodal knowledge graph representations}, \bibinfo{year}{2022}

\bibitem{chen2022multi}
\bibinfo{author}{Chen S}, \bibinfo{author}{Li B}.
\newblock \bibinfo{title}{Multi-modal dynamic graph transformer for visual grounding}.
\newblock In: \bibinfo{booktitle}{Proceedings of the IEEE/CVF Conference on Computer Vision and Pattern Recognition}, \bibinfo{year}{2022}.
\newblock \bibinfo{pages}{15534--15543}

\bibitem{edge2025localglobalgraphrag}
\bibinfo{author}{Edge D}, \bibinfo{author}{Trinh H}, \bibinfo{author}{Cheng N}, et~al.
\newblock \bibinfo{title}{From local to global: A graph rag approach to query-focused summarization}, \bibinfo{year}{2025}

\bibitem{creswell2023selectioninference}
\bibinfo{author}{Creswell A}, \bibinfo{author}{Shanahan M}, \bibinfo{author}{Higgins I}.
\newblock \bibinfo{title}{Selection-inference: Exploiting large language models for interpretable logical reasoning}.
\newblock In: \bibinfo{booktitle}{The Eleventh International Conference on Learning Representations}, \bibinfo{year}{2023}

\bibitem{wei2023symbol}
\bibinfo{author}{Wei J}, \bibinfo{author}{Hou L}, \bibinfo{author}{Lampinen A~K}, et~al.
\newblock \bibinfo{title}{Symbol tuning improves in-context learning in language models}.
\newblock In: \bibinfo{booktitle}{The 2023 Conference on Empirical Methods in Natural Language Processing}, \bibinfo{year}{2023}

\bibitem{talmor2021multimodalqa}
\bibinfo{author}{Talmor A}, \bibinfo{author}{Yoran O}, \bibinfo{author}{Catav A}, et~al.
\newblock \bibinfo{title}{Multimodal{\{}qa{\}}: complex question answering over text, tables and images}.
\newblock In: \bibinfo{booktitle}{International Conference on Learning Representations}, \bibinfo{year}{2021}

\bibitem{zhang2023multimodal}
\bibinfo{author}{Zhang N}, \bibinfo{author}{Li L}, \bibinfo{author}{Chen X}, et~al.
\newblock \bibinfo{title}{Multimodal analogical reasoning over knowledge graphs}.
\newblock In: \bibinfo{booktitle}{The Eleventh International Conference on Learning Representations}, \bibinfo{year}{2023}

\bibitem{guo2024can}
\bibinfo{author}{Guo D}, \bibinfo{author}{Cao C}, \bibinfo{author}{Yuan F}, et~al.
\newblock \bibinfo{title}{Can multimodal large language model think analogically?}
\newblock \bibinfo{journal}{arXiv preprint arXiv:2411.01307}, \bibinfo{year}{2024}

\bibitem{radford2021learning}
\bibinfo{author}{Radford A}, \bibinfo{author}{Kim J~W}, \bibinfo{author}{Hallacy C}, et~al.
\newblock \bibinfo{title}{Learning transferable visual models from natural language supervision}.
\newblock In: \bibinfo{booktitle}{International conference on machine learning}. \bibinfo{organization}{PmLR}\bibinfo{year}{2021}.
\newblock \bibinfo{pages}{8748--8763}

\bibitem{chen2024unified}
\bibinfo{author}{Chen X}, \bibinfo{author}{Wang C}, \bibinfo{author}{Xue Y}, et~al.
\newblock \bibinfo{title}{Unified hallucination detection for multimodal large language models}.
\newblock In: \bibinfo{booktitle}{Proceedings of the 62nd Annual Meeting of the Association for Computational Linguistics (Volume 1: Long Papers)}. \bibinfo{publisher}{Association for Computational Linguistics}\bibinfo{year}{2024}.
\newblock \bibinfo{pages}{3235--3252}

\bibitem{liu2024towards}
\bibinfo{author}{Liu X}, \bibinfo{author}{LI R}, \bibinfo{author}{Ji W}, et~al.
\newblock \bibinfo{title}{Towards robust multi-modal reasoning via model selection}.
\newblock In: \bibinfo{booktitle}{The Twelfth International Conference on Learning Representations}, \bibinfo{year}{2024}

\bibitem{choi2024multimodal}
\bibinfo{author}{Choi I}, \bibinfo{author}{Yun S}, \bibinfo{author}{Xin J}, et~al.
\newblock \bibinfo{title}{Multimodal graph-llm: Leveraging graph-enhanced llms for multimodal healthcare predictions}, \bibinfo{year}{2024}

\bibitem{chen2022hybrid}
\bibinfo{author}{Chen X}, \bibinfo{author}{Zhang N}, \bibinfo{author}{Li L}, et~al.
\newblock \bibinfo{title}{Hybrid transformer with multi-level fusion for multimodal knowledge graph completion}.
\newblock In: \bibinfo{booktitle}{{SIGIR} '22: The 45th International {ACM} {SIGIR} Conference on Research and Development in Information Retrieval, Madrid, Spain, July 11 - 15, 2022}. \bibinfo{publisher}{{ACM}}\bibinfo{year}{2022}.
\newblock \bibinfo{pages}{904--915}

\bibitem{Kaiser2017}
\bibinfo{author}{Kaiser L}, \bibinfo{author}{Gomez A~N}, \bibinfo{author}{Shazeer N}, et~al.
\newblock \bibinfo{title}{One model to learn them all}.
\newblock \bibinfo{journal}{CoRR}, \bibinfo{year}{2017}, \bibinfo{volume}{abs/1706.05137}.
\newblock \bibinfo{note}{ArXiv:1706.05137}

\bibitem{Zaheer2020}
\bibinfo{author}{Zaheer M}, \bibinfo{author}{Guruganesh G}, \bibinfo{author}{et~al}.
\newblock \bibinfo{title}{Big bird: Transformers for longer sequences}.
\newblock In: \bibinfo{booktitle}{Advances in Neural Information Processing Systems (NeurIPS)}, \bibinfo{year}{2020}, volume~\bibinfo{volume}{33}.
\newblock \bibinfo{pages}{17283--17297}

\bibitem{Jaegle2021}
\bibinfo{author}{Jaegle A}, \bibinfo{author}{Gimeno F}, \bibinfo{author}{et~al}.
\newblock \bibinfo{title}{Perceiver: General perception with iterative attention}.
\newblock In: \bibinfo{booktitle}{Proceedings of the 38th International Conference on Machine Learning (ICML)}, \bibinfo{year}{2021}.
\newblock \bibinfo{pages}{4651--4664}

\bibitem{Tay2020}
\bibinfo{author}{Tay Y}, \bibinfo{author}{Dehghani M}, \bibinfo{author}{Bahri D}, et~al.
\newblock \bibinfo{title}{Efficient transformers: A survey}.
\newblock \bibinfo{journal}{arXiv preprint arXiv:2009.06732}, \bibinfo{year}{2020}

\bibitem{Lewis2020}
\bibinfo{author}{Lewis P}, \bibinfo{author}{Perez E}, \bibinfo{author}{Piktus A}, et~al.
\newblock \bibinfo{title}{Retrieval-augmented generation for knowledge-intensive nlp tasks}.
\newblock In: \bibinfo{booktitle}{Advances in Neural Information Processing Systems (NeurIPS)}, \bibinfo{year}{2020}, volume~\bibinfo{volume}{33}.
\newblock \bibinfo{pages}{9459--9474}

\bibitem{Hamilton2017}
\bibinfo{author}{Hamilton W~L}, \bibinfo{author}{Ying R}, \bibinfo{author}{Leskovec J}.
\newblock \bibinfo{title}{Inductive representation learning on large graphs}.
\newblock In: \bibinfo{booktitle}{Advances in Neural Information Processing Systems (NeurIPS)}, \bibinfo{year}{2017}, volume~\bibinfo{volume}{30}.
\newblock \bibinfo{pages}{1024--1034}

\bibitem{Zeng2020}
\bibinfo{author}{Zeng H}, \bibinfo{author}{Zhou H}, \bibinfo{author}{Srivastava A}, et~al.
\newblock \bibinfo{title}{Graphsaint: Graph sampling based inductive learning method}.
\newblock In: \bibinfo{booktitle}{Proc. of International Conference on Learning Representations (ICLR)}, \bibinfo{year}{2020}

\bibitem{Ying2018}
\bibinfo{author}{Ying R}, \bibinfo{author}{You J}, \bibinfo{author}{Morris C}, et~al.
\newblock \bibinfo{title}{Hierarchical graph representation learning with differentiable pooling}.
\newblock In: \bibinfo{booktitle}{Advances in Neural Information Processing Systems (NeurIPS)}, \bibinfo{year}{2018}, volume~\bibinfo{volume}{31}.
\newblock \bibinfo{pages}{4800--4810}

\bibitem{Zheng2021}
\bibinfo{author}{Zheng D}, \bibinfo{author}{Song X}, \bibinfo{author}{Yang C}, et~al.
\newblock \bibinfo{title}{Distributed hybrid cpu and gpu training for graph neural networks on billion-scale graphs}.
\newblock \bibinfo{journal}{arXiv preprint arXiv:2010.05337}, \bibinfo{year}{2021}

\bibitem{Bengio2009}
\bibinfo{author}{Bengio Y}, \bibinfo{author}{Louradour J}, \bibinfo{author}{Collobert R}, et~al.
\newblock \bibinfo{title}{Curriculum learning}.
\newblock In: \bibinfo{booktitle}{Proceedings of the 26th International Conference on Machine Learning (ICML)}, \bibinfo{year}{2009}.
\newblock \bibinfo{pages}{41--48}

\bibitem{Hinton2015}
\bibinfo{author}{Hinton G}, \bibinfo{author}{Vinyals O}, \bibinfo{author}{Dean J}.
\newblock \bibinfo{title}{Distilling the knowledge in a neural network}.
\newblock \bibinfo{journal}{CoRR}, \bibinfo{year}{2015}, \bibinfo{volume}{abs/1503.02531}

\bibitem{Han2016}
\bibinfo{author}{Han S}, \bibinfo{author}{Mao H}, \bibinfo{author}{Dally W~J}.
\newblock \bibinfo{title}{Deep compression: Compressing deep neural networks with pruning, trained quantization and huffman coding}.
\newblock In: \bibinfo{booktitle}{Proceedings of the 4th International Conference on Learning Representations (ICLR)}, \bibinfo{year}{2016}

\bibitem{Ying2021}
\bibinfo{author}{Ying C}, \bibinfo{author}{Cai T}, \bibinfo{author}{Luo S}, et~al.
\newblock \bibinfo{title}{Do transformers really perform bad for graph representation?}
\newblock In: \bibinfo{booktitle}{Advances in Neural Information Processing Systems (NeurIPS)}, \bibinfo{year}{2021}, volume~\bibinfo{volume}{34}.
\newblock \bibinfo{pages}{28877--28888}

\bibitem{Kang2017}
\bibinfo{author}{Kang Y}, \bibinfo{author}{Hauswald J}, \bibinfo{author}{Gao C}, et~al.
\newblock \bibinfo{title}{Neurosurgeon: Collaborative intelligence between the cloud and mobile edge}.
\newblock In: \bibinfo{booktitle}{Proceedings of the 22nd International Conference on Architectural Support for Programming Languages and Operating Systems (ASPLOS)}. \bibinfo{publisher}{ACM}\bibinfo{year}{2017}.
\newblock \bibinfo{pages}{615--629}

\bibitem{TAO2025104353}
\bibinfo{author}{Tao Y}, \bibinfo{author}{Liu W}, \bibinfo{author}{Chen J}, et~al.
\newblock \bibinfo{title}{A graph-based multimodal data fusion framework for identifying urban functional zone}.
\newblock \bibinfo{journal}{International Journal of Applied Earth Observation and Geoinformation}, \bibinfo{year}{2025}, \bibinfo{volume}{136}: \bibinfo{pages}{104353}

\bibitem{hoadley2018cell}
\bibinfo{author}{Hoadley K~A}, \bibinfo{author}{Yau C}, \bibinfo{author}{Hinoue T}, et~al.
\newblock \bibinfo{title}{Cell-of-origin patterns dominate the molecular classification of 10,000 tumors from 33 types of cancer}.
\newblock \bibinfo{journal}{Cell}, \bibinfo{year}{2018}, \bibinfo{volume}{173}: \bibinfo{pages}{291--304}

\bibitem{cancer2012comprehensive}
\bibinfo{author}{13 B~~W~H~~H~M~S~C~L~~~P~P~J~~K~R}, \bibinfo{author}{data analysis: Baylor College~of Medicine Creighton Chad J 22 23 Donehower Lawrence A 22 23 24~25 G}, \bibinfo{author}{for Systems Biology Reynolds Sheila 31 Kreisberg Richard B 31 Bernard Brady 31 Bressler Ryan 31 Erkkila Timo 32 Lin Jake 31 Thorsson Vesteinn 31 Zhang Wei 33 Shmulevich Ilya~31 I}, et~al.
\newblock \bibinfo{title}{Comprehensive molecular portraits of human breast tumours}.
\newblock \bibinfo{journal}{Nature}, \bibinfo{year}{2012}, \bibinfo{volume}{490}: \bibinfo{pages}{61--70}

\bibitem{Cai_Wang_Li_Zhang_Zhu_2024}
\bibinfo{author}{Cai J}, \bibinfo{author}{Wang X}, \bibinfo{author}{Li H}, et~al.
\newblock \bibinfo{title}{Multimodal graph neural architecture search under distribution shifts}.
\newblock \bibinfo{journal}{Proceedings of the AAAI Conference on Artificial Intelligence}, \bibinfo{year}{2024}, \bibinfo{volume}{38}: \bibinfo{pages}{8227--8235}

\bibitem{saifuddin2025hypergclmultimodalgraphcontrastive}
\bibinfo{author}{Saifuddin K~M}, \bibinfo{author}{Ji S}, \bibinfo{author}{Akbas E}.
\newblock \bibinfo{title}{Hypergcl: Multi-modal graph contrastive learning via learnable hypergraph views}, \bibinfo{year}{2025}

\bibitem{lee-etal-2023-vista}
\bibinfo{author}{Lee J}, \bibinfo{author}{Chung C}, \bibinfo{author}{Lee H}, et~al.
\newblock \bibinfo{title}{Vista: Visual-textual knowledge graph representation learning}.
\newblock In: \bibinfo{booktitle}{Findings of the Association for Computational Linguistics: EMNLP 2023}. \bibinfo{publisher}{Association for Computational Linguistics}\bibinfo{year}{2023}.
\newblock \bibinfo{pages}{7314--7328}

\bibitem{10.1145/3581783.3612266}
\bibinfo{author}{Wang X}, \bibinfo{author}{Meng B}, \bibinfo{author}{Chen H}, et~al.
\newblock \bibinfo{title}{Tiva-kg: A multimodal knowledge graph with text, image, video and audio}.
\newblock In: \bibinfo{booktitle}{Proceedings of the 31st ACM International Conference on Multimedia}. \bibinfo{publisher}{Association for Computing Machinery}\bibinfo{year}{2023}, MM '23.
\newblock \bibinfo{pages}{2391–2399}

\bibitem{ijcai2022p482}
\bibinfo{author}{Tian Y}, \bibinfo{author}{Zhang C}, \bibinfo{author}{Guo Z}, et~al.
\newblock \bibinfo{title}{Recipe2vec: Multi-modal recipe representation learning with graph neural networks}.
\newblock In: \bibinfo{booktitle}{Proceedings of the Thirty-First International Joint Conference on Artificial Intelligence, {IJCAI-22}}. \bibinfo{publisher}{International Joint Conferences on Artificial Intelligence Organization}\bibinfo{year}{2022}.
\newblock \bibinfo{pages}{3473--3479}.
\newblock \bibinfo{note}{Main Track}

\bibitem{Johnson_2017_CVPR}
\bibinfo{author}{Johnson J}, \bibinfo{author}{Hariharan B}, \bibinfo{author}{van~der Maaten L}, et~al.
\newblock \bibinfo{title}{Clevr: A diagnostic dataset for compositional language and elementary visual reasoning}.
\newblock In: \bibinfo{booktitle}{Proceedings of the IEEE Conference on Computer Vision and Pattern Recognition (CVPR)}, \bibinfo{year}{2017}

\bibitem{alonso2025visionlanguagemodelsstrugglealign}
\bibinfo{author}{Alonso I}, \bibinfo{author}{Salaberria A}, \bibinfo{author}{Azkune G}, et~al.
\newblock \bibinfo{title}{Vision-language models struggle to align entities across modalities}, \bibinfo{year}{2025}

\bibitem{mousselly2018multimodal}
\bibinfo{author}{Mousselly-Sergieh H}, \bibinfo{author}{Botschen T}, \bibinfo{author}{Gurevych I}, et~al.
\newblock \bibinfo{title}{A multimodal translation-based approach for knowledge graph representation learning}.
\newblock In: \bibinfo{booktitle}{Proceedings of the Seventh Joint Conference on Lexical and Computational Semantics}, \bibinfo{year}{2018}.
\newblock \bibinfo{pages}{225--234}

\bibitem{shirai-etal-2022-visual}
\bibinfo{author}{Shirai K}, \bibinfo{author}{Hashimoto A}, \bibinfo{author}{Nishimura T}, et~al.
\newblock \bibinfo{title}{Visual recipe flow: A dataset for learning visual state changes of objects with recipe flows}.
\newblock In: \bibinfo{booktitle}{Proceedings of the 29th International Conference on Computational Linguistics}. \bibinfo{publisher}{International Committee on Computational Linguistics}\bibinfo{year}{2022}.
\newblock \bibinfo{pages}{3570--3577}

\bibitem{lin-etal-2020-recipe}
\bibinfo{author}{Lin A}, \bibinfo{author}{Rao S}, \bibinfo{author}{Celikyilmaz A}, et~al.
\newblock \bibinfo{title}{A recipe for creating multimodal aligned datasets for sequential tasks}.
\newblock In: \bibinfo{booktitle}{Proceedings of the 58th Annual Meeting of the Association for Computational Linguistics}. \bibinfo{publisher}{Association for Computational Linguistics}\bibinfo{year}{2020}.
\newblock \bibinfo{pages}{4871--4884}

\bibitem{fang2025graphgpt}
\bibinfo{author}{Fang Y}, \bibinfo{author}{Jin B}, \bibinfo{author}{Shen J}, et~al.
\newblock \bibinfo{title}{Graphgpt-o: Synergistic multimodal comprehension and generation on graphs}.
\newblock \bibinfo{journal}{arXiv preprint arXiv:2502.11925}, \bibinfo{year}{2025}

\bibitem{jin2024instructg2i}
\bibinfo{author}{Jin B}, \bibinfo{author}{Pang Z}, \bibinfo{author}{Guo B}, et~al.
\newblock \bibinfo{title}{Instructg2i: Synthesizing images from multimodal attributed graphs}.
\newblock \bibinfo{journal}{arXiv preprint arXiv:2410.07157}, \bibinfo{year}{2024}

\bibitem{kipf2017semi}
\bibinfo{author}{Kipf T~N}, \bibinfo{author}{Welling M}.
\newblock \bibinfo{title}{Semi-supervised classification with graph convolutional networks}.
\newblock In: \bibinfo{booktitle}{International Conference on Learning Representations (ICLR)}, \bibinfo{year}{2017}

\bibitem{velickovic2018graph}
\bibinfo{author}{Veli{\v{c}}kovi{'c} P}, \bibinfo{author}{Cucurull G}, \bibinfo{author}{Casanova A}, et~al.
\newblock \bibinfo{title}{Graph attention networks}.
\newblock In: \bibinfo{booktitle}{International Conference on Learning Representations (ICLR)}, \bibinfo{year}{2018}

\bibitem{xu2019powerful}
\bibinfo{author}{Xu K}, \bibinfo{author}{Hu W}, \bibinfo{author}{Leskovec J}, et~al.
\newblock \bibinfo{title}{How powerful are graph neural networks?}
\newblock In: \bibinfo{booktitle}{International Conference on Learning Representations (ICLR)}, \bibinfo{year}{2019}

\bibitem{zhang2023gnnacc}
\bibinfo{author}{Zhang S}, \bibinfo{author}{Sohrabizadeh A}, \bibinfo{author}{Wan C}, et~al.
\newblock \bibinfo{title}{A survey on graph neural network acceleration: Algorithms, systems, and customized hardware}.
\newblock \bibinfo{journal}{arXiv preprint arXiv:2306.14052}, \bibinfo{year}{2023}

\bibitem{dai2024trustworthy}
\bibinfo{author}{Dai E}, \bibinfo{author}{Zhao T}, \bibinfo{author}{Zhu H}, et~al.
\newblock \bibinfo{title}{A comprehensive survey on trustworthy graph neural networks: Privacy, robustness, fairness, and explainability}.
\newblock \bibinfo{journal}{Machine Intelligence Research}, \bibinfo{year}{2024}, \bibinfo{volume}{21}: \bibinfo{pages}{1011–1061}

\bibitem{han2025graphrag}
\bibinfo{author}{Han H}, \bibinfo{author}{Wang Y}, \bibinfo{author}{Shomer H}, et~al.
\newblock \bibinfo{title}{Retrieval-augmented generation with graphs (graphrag)}.
\newblock \bibinfo{journal}{arXiv preprint arXiv:2501.00309}, \bibinfo{year}{2025}

\bibitem{openai2023gpt4}
\bibinfo{author}{OpenAI}.
\newblock \bibinfo{title}{Gpt-4 technical report}.
\newblock \bibinfo{journal}{arXiv preprint arXiv:2303.08774}, \bibinfo{year}{2023}

\bibitem{grattafiori2024llama}
\bibinfo{author}{Grattafiori A}, \bibinfo{author}{Dubey A}, \bibinfo{author}{Jauhri A}, et~al.
\newblock \bibinfo{title}{The llama 3 herd of models}.
\newblock \bibinfo{journal}{arXiv preprint arXiv:2407.21783}, \bibinfo{year}{2024}

\bibitem{yang2025qwen3}
\bibinfo{author}{Yang A}, \bibinfo{author}{Li A}, \bibinfo{author}{Yang B}, et~al.
\newblock \bibinfo{title}{Qwen3 technical report}.
\newblock \bibinfo{journal}{arXiv preprint arXiv:2505.09388}, \bibinfo{year}{2025}

\bibitem{vaswani2017attention}
\bibinfo{author}{Vaswani A}, \bibinfo{author}{Shazeer N}, \bibinfo{author}{Parmar N}, et~al.
\newblock \bibinfo{title}{Attention is all you need}.
\newblock \bibinfo{journal}{Advances in neural information processing systems}, \bibinfo{year}{2017}, \bibinfo{volume}{30}

\bibitem{openai2023gpt4vsystemcard}
\bibinfo{author}{OpenAI}.
\newblock \bibinfo{title}{Gpt-4v(ision) system card}, \bibinfo{year}{2023}.
\newblock \bibinfo{note}{Accessed: 2025-08-31}

\bibitem{bai2025qwen25vl}
\bibinfo{author}{Bai S}, \bibinfo{author}{Chen K}, \bibinfo{author}{Liu X}, et~al.
\newblock \bibinfo{title}{Qwen2. 5-vl technical report}.
\newblock \bibinfo{journal}{arXiv preprint arXiv:2502.13923}, \bibinfo{year}{2025}

\bibitem{wang2025internvl35}
\bibinfo{author}{Wang W}, \bibinfo{author}{Gao Z}, \bibinfo{author}{Gu L}, et~al.
\newblock \bibinfo{title}{Internvl3.5: Advancing open-source multimodal models in versatility, reasoning, and efficiency}, \bibinfo{year}{2025}

\bibitem{vteam2025glm45v}
\bibinfo{author}{Team V}, \bibinfo{author}{Hong W}, \bibinfo{author}{Yu W}, et~al.
\newblock \bibinfo{title}{Glm-4.5v and glm-4.1v-thinking: Towards versatile multimodal reasoning with scalable reinforcement learning}, \bibinfo{year}{2025}

\bibitem{hurst2024gpt}
\bibinfo{author}{Hurst A}, \bibinfo{author}{Lerer A}, \bibinfo{author}{Goucher A~P}, et~al.
\newblock \bibinfo{title}{Gpt-4o system card}.
\newblock \bibinfo{journal}{arXiv preprint arXiv:2410.21276}, \bibinfo{year}{2024}

\bibitem{comanici2025gemini}
\bibinfo{author}{Comanici G}, \bibinfo{author}{Bieber E}, \bibinfo{author}{Schaekermann M}, et~al.
\newblock \bibinfo{title}{Gemini 2.5: Pushing the frontier with advanced reasoning, multimodality, long context, and next generation agentic capabilities}.
\newblock \bibinfo{journal}{arXiv preprint arXiv:2507.06261}, \bibinfo{year}{2025}

\bibitem{han2024onellm}
\bibinfo{author}{Han J}, \bibinfo{author}{Gong K}, \bibinfo{author}{Zhang Y}, et~al.
\newblock \bibinfo{title}{Onellm: One framework to align all modalities with language}.
\newblock In: \bibinfo{booktitle}{Proceedings of the IEEE/CVF Conference on Computer Vision and Pattern Recognition}, \bibinfo{year}{2024}.
\newblock \bibinfo{pages}{26584--26595}

\bibitem{tang2024codi}
\bibinfo{author}{Tang Z}, \bibinfo{author}{Yang Z}, \bibinfo{author}{Khademi M}, et~al.
\newblock \bibinfo{title}{Codi-2: In-context interleaved and interactive any-to-any generation}.
\newblock In: \bibinfo{booktitle}{Proceedings of the IEEE/CVF Conference on Computer Vision and Pattern Recognition}, \bibinfo{year}{2024}.
\newblock \bibinfo{pages}{27425--27434}

\bibitem{team2025kimi}
\bibinfo{author}{Team K}, \bibinfo{author}{Bai Y}, \bibinfo{author}{Bao Y}, et~al.
\newblock \bibinfo{title}{Kimi k2: Open agentic intelligence}.
\newblock \bibinfo{journal}{arXiv preprint arXiv:2507.20534}, \bibinfo{year}{2025}

\bibitem{zeng2025glm}
\bibinfo{author}{Zeng A}, \bibinfo{author}{Lv X}, \bibinfo{author}{Zheng Q}, et~al.
\newblock \bibinfo{title}{Glm-4.5: Agentic, reasoning, and coding (arc) foundation models}.
\newblock \bibinfo{journal}{arXiv preprint arXiv:2508.06471}, \bibinfo{year}{2025}

\end{thebibliography}

\end{document}